%% file: main.tex
\documentclass[lettersize,journal]{IEEEtran}
\usepackage{amsmath,amsfonts}
\usepackage{algorithmic}
\usepackage{algorithm}
\usepackage{array}
\usepackage{textcomp}
\usepackage{stfloats}
\usepackage{url}
\usepackage{verbatim}
\usepackage{graphicx}
\usepackage{cite}
\usepackage{xspace}
\usepackage{hyperref}     
\usepackage[numbers]{natbib}
\usepackage{booktabs}   
\usepackage{multirow}       
\usepackage[table]{xcolor} 

\usepackage{caption}
\usepackage{subfigure}
\usepackage{lipsum}  
\usepackage{tcolorbox}  
\usepackage{lipsum}

\usepackage{pgf}
\hypersetup{
    colorlinks=true,
    linkcolor=green,
    filecolor=magenta,      
    urlcolor=cyan,
    citecolor=cyan
} 

\definecolor{lightblue}{RGB}{0,225,225}
\definecolor{mygreen}{RGB}{166,219,162}
\definecolor{myblue}{RGB}{0,0,150}
\definecolor{lygblue}{RGB}{0, 123, 255}
\definecolor{cyqpurple}{RGB}{128, 0, 128} 
\newcommand{\numComma}[1]{%
    \pgfmathparse{#1}%
    \pgfmathprintnumber[fixed, fixed zerofill, precision=0]{\pgfmathresult}%
}

\newcommand{\ourmodel}{CDBN\xspace}

\begin{document}

\title{Data-Efficient CLIP-Powered Dual-Branch Networks for Source-Free Unsupervised Domain Adaptation\thanks{This work has been submitted to the IEEE for possible publication. Copyright may be transferred without notice, after which this version may no longer be accessible.}}

\author{
  \begin{tabular}{c} 
    Yongguang Li$^{1,3,\dag}$\thanks{\dag: Equal Contribution.} \and Yueqi Cao$^{2,3,\dag}$ \and Jindong Li$^{4}$  \and Qi Wang$^{4,5}$ \and Shengsheng Wang$^{2,3,*}$\thanks{*: Corresponding Author.}
  \end{tabular} \\
  \begin{tabular}{c}
    $^{1}$College of Software, Jilin University, Changchun 130012, China \\
    $^{2}$College of Computer Science and Technology, Jilin University, Changchun 130012, China \\
    $^{3}$Key Laboratory of Symbolic Computation and Knowledge Engineering of Ministry of Education,\\ Jilin University, Changchun 130012, China\\
    $^{4}$School of Artificial Intelligence, Jilin University, Changchun 130012, China \\ 
    $^{5}$Engineering Research Center of Knowledge-Driven Human-Machine Intelligence, Ministry of Education, \\ Jilin University, Changchun 130012, China\\ 
    \{liyg22, caoyq22, jdli21\}@mails.jlu.edu.cn, \{qiwang,~wss\}@jlu.edu.cn
  \end{tabular}
}

\maketitle
\thispagestyle{plain} 
\begin{abstract}

Source-free Unsupervised Domain Adaptation (SF-UDA) aims to transfer a model's performance from a labeled source domain to an unlabeled target domain without direct access to source samples, addressing critical data privacy concerns. However, most existing SF-UDA approaches assume the availability of abundant source domain samples, which is often impractical due to the high cost of data annotation. To address the dual challenges of limited source data and privacy concerns, we introduce a data-efficient, CLIP-powered dual-branch network (CDBN). This architecture consists of a cross-domain feature transfer branch and a target-specific feature learning branch, leveraging high-confidence target domain samples to transfer text features of source domain categories while learning target-specific soft prompts. By fusing the outputs of both branches, our approach not only effectively transfers source domain category semantic information to the target domain but also reduces the negative impacts of noise and domain gaps during target training. Furthermore, we propose an unsupervised optimization strategy driven by accurate classification and diversity, preserving the classification capability learned from the source domain while generating more confident and diverse predictions in the target domain. CDBN achieves near state-of-the-art performance with far fewer source domain samples than existing methods across 31 transfer tasks on seven datasets.

\end{abstract}


\section{Introduction}
\label{sec_Introduction}

Deep learning has achieved remarkable performance in various computer vision tasks \cite{2024_TCSVT_PUBCLIP, 2023_ICCV_DETR, 2023_CVPR_InternImage}. However, these achievements largely rely on the assumption that the training and test data are independently and identically distributed (i.i.d.) \cite{learningdataset}.
In real-world scenarios, differences in data collection conditions, environmental factors, and other variables often cause domain shifts between the training data (source domain) and test data (target domain),significantly reducing model performance.
To address this issue, previous works \cite{2017_PMLR_JAN, 2015_ICML_Unsupervised-Domain-Adaptation-by_Backpropogation} have proposed unsupervised domain adaptation (UDA), employing adversarial training \cite{2022_TCSVT_DAFL,2015_ICML_Unsupervised-Domain-Adaptation-by_Backpropogation}, metric learning \cite{2017_PMLR_JAN}, and other techniques to learn domain-invariant features, thereby facilitating the transfer of models trained on the source domain to the unlabeled target domain. 

Despite the effectiveness of UDA methods in addressing domain shifts, they typically assume that the source domain is accessible during target domain training and contains sufficient labeled data for training. However, this assumption is often unrealistic in practical settings due to the high cost of data annotation and the need to protect data privacy.
For instance, in the diabetic retinopathy dataset used in clinical research \cite{2016_JAMA_Development-and-Validation}, each image is annotated by 7 to 8 board-certified ophthalmologists, significantly increasing the cost of labeling \cite{2021_CVPR_PCS_Prototypical-Cross-Domain-Self-Supervised-Learning}. 
Additionally, in fields such as medicine and remote sensing \cite{2021_Remote-Sensing_Grand-Challenges}, labeled data often cannot be distributed or reused due to privacy protection and commercial copyright concerns.

\begin{figure*}[t]
    \centering
    \includegraphics[width=0.99\linewidth]{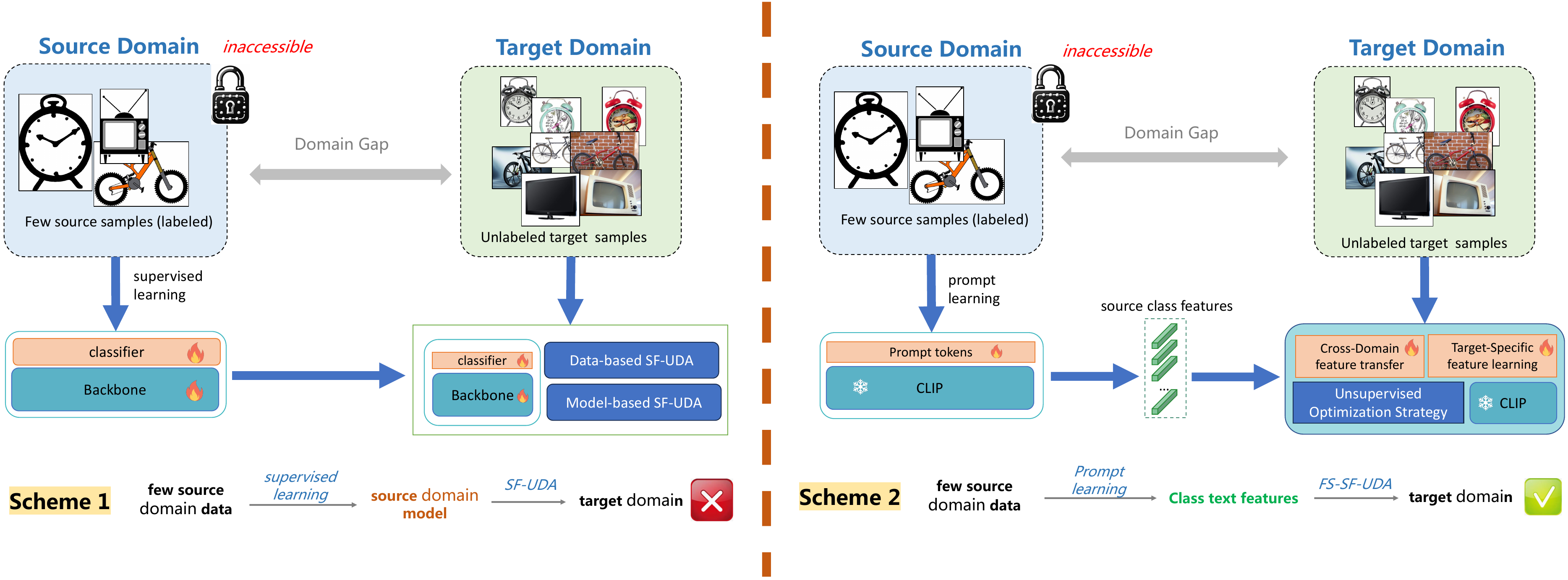}
    \caption{Scheme 1 illustrates the issue with previous SF-UDA methods, where using a fully trained model from a source domain as an information source for a target domain leads to poor performance when source domain samples are limited. To address this issue, Scheme 2 presents our proposed solution, which involves transferring category feature representations obtained from a source-domain-based prompt learning method to the target domain, effectively resolving the SF-UDA problem in scenarios with few source domain samples.}
    \label{fig: fig_1}
\end{figure*}

Researchers have proposed two types of methods to address the challenges of restricted access to the source domain and the limited number of source domain samples.
Source-free unsupervised domain adaptation (SF-UDA) methods \cite{2020_PMLR_SHOT,2022_TCSVT_DAFL,2023_TCSVT_dcl} address the issue of inaccessible source data by enabling model transfer to the target domain without requiring direct access to the source domain. Meanwhile, few-shot unsupervised domain adaptation (FS-UDA) methods \cite{2021_CVPR_PCS_Prototypical-Cross-Domain-Self-Supervised-Learning, 2020_arXiv_CDS_Cross-Domain-Self-Supervised-Learning, 2024_ICIC_PMPA_Few-Shot-Domain-Adaptation-via-Prompt-Guided} tackle the problem of limited labeled data by facilitating effective model transfer with minimal source samples. However, methods that address the dual challenges of restricted access to source domain data and small amounts of source domain samples, known as SF-UDA methods under few-shot source domain scenarios (replaced by FS-SF-UDA in this paper), have yet to be fully explored.

In recent years, prompt learning for CLIP models \cite{2023_CVPR_MaPLe,2022_IJCV_CoOp,2024_AAAI_LAMM} has demonstrated significant advantages in few-shot classification tasks. These methods typically leverage CLIP as a backbone network and employ prompt learning techniques to achieve strong downstream task performance, even with only a limited number of samples per class. Inspired by this, we propose a data-efficient \textbf{\underline{C}}LIP-powered \textbf{\underline{D}}ual-\textbf{\underline{B}}ranch \textbf{\underline{N}}etwork (dubbed as CDBN) for FS-SF-UDA.

In scenarios with only a few samples in the source domain, our goal is to efficiently extract useful category semantic information from the source domain and deliver it to the target domain as a reference for unsupervised training. Specifically, as shown in Scheme 1 of Figure \ref{fig: fig_1}, traditional methods that rely on supervised training of a source domain model perform poorly when the number of samples in the source domain is limited. To address this issue, as illustrated in Scheme 2 of Figure \ref{fig: fig_1}, we employ a CLIP-based prompt learning approach to learn category text features with a certain degree of generalization from a small number of source domain samples. We then leverage these category text features as carriers of source domain semantic information to guide the unsupervised training process in the target domain.

However, the combination of limited source domain samples and the domain gap presents a challenge, as it introduces noise and domain shift into the transferred category text features.
To address this, we designed a cross-modal dual-branch network consisting of a cross-domain feature transfer branch and a target-specific feature learning branch. The cross-domain feature transfer branch utilizes high-confidence samples from the target domain to transfer the learned source domain category text features, reducing the impact of domain discrepancy. Meanwhile, the target-specific feature learning branch focuses on learning independent category features from the target domain, avoiding over-reliance on noisy source domain text features. The synergy between these two branches effectively enhances the feature representation capabilities in the target domain.

Furthermore, since there are no labels during the training process in the target domain, we propose an unsupervised optimization strategy driven by accurate classification and diversity. This strategy includes three core loss functions: first, the cross-entropy loss from high-confidence target domain samples derived from source domain category text features ensures that the model retains the classification ability learned from the source domain; second, the multi-view consistency loss encourages the model to extract consistent feature representations from different perspectives, enhancing the generalization capability of the learned category feature distribution; finally, the mutual information maximization loss between inputs and outputs promotes diversity in the category distribution, preventing the model from overfitting to simple categories. Together, these three components ensure that the model can learn target domain features accurately and robustly under unsupervised conditions.

We conducted experiments on 31 transfer tasks across 7 publicly available domain adaptation datasets. Despite the dual challenges of limited source samples and restricted access to source domain data, CDBN achieved state-of-the-art performance on multiple datasets and showed competitive results comparable to the best existing methods on others. These outcomes demonstrate the robustness and effectiveness of our CDBN approach in addressing the FS-SF-UDA problem.

Our main contributions could be summarized as follows:
\begin{itemize}

    \item We propose a data-efficient CLIP-powered dual-branch network for Few-Shot Source-Free Unsupervised Domain Adaptation, addressing the dual challenges of limited source domain samples and restricted access to source domain data.

    \item We introduce a cross-modal dual-branch network structure that effectively leverages source domain category semantic information during the unsupervised fine-tuning process in the target domain while mitigating the effects of source domain noise and domain discrepancy. Additionally, it combines the strengths of both branches, significantly enhancing performance in the target domain.

    \item We introduce an unsupervised optimization strategy that balances accurate classification with feature diversity. This strategy retains the classification capabilities learned from the source domain while promoting precise and diverse classification results in the target domain, thereby improving the model’s performance on the target domain.

    \item We conduct extensive experiments across 31 transfer tasks on 7 public datasets, demonstrating our \ourmodel achieves state-of-the-art performance over existing methods.
\end{itemize}

\section{Related Work}
\label{sec_Realted Work}

\subsection{Source-Free Unsupervised Domain Adaptation}
Source-Free Unsupervised Domain Adaptation (SF-UDA) aims to address the challenge in unsupervised domain adaptation where source domain data is inaccessible during training on the target domain. Most previous approaches can be categorized into two main types: data-based and model-based methods. Data-based methods typically generate virtual domains from the source domain model to replace the source data during training on the target domain, while model-based methods rely on self-supervised training directly on the source domain model to enhance its performance on the target domain after adaptation.
For example, \citet{2021_TCSVT_VDM-DA} proposes utilizing a model trained on the source domain to generate a virtual domain in the feature space based on an approximate Gaussian mixture model. By progressively increasing the compactness of the target domain features, this approach reduces the domain gap between the virtual and target domains, thus improving the model's performance on the target domain. Similarly, \citet{2022_ECCV_BMD} introduces a multi-center clustering strategy within each class along with a dynamic pseudo-labeling strategy. By incorporating network updates during model adaptation, this method significantly enhances the performance of several representative SF-UDA approaches \cite{2020_PMLR_SHOT,2021_ICCV_G-SFDA}.

\subsection{Few-Shot Unsupervised Domain Adaptation}
Few-Shot Unsupervised Domain Adaptation (FS-UDA) aims to address the challenge of having limited labeled samples in the source domain during training on the target domain, even though source domain samples are accessible. Previous methods have explored different scenarios to tackle this issue. For instance, \citet{2021_CVPR_PCS_Prototypical-Cross-Domain-Self-Supervised-Learning} proposes a method to address the problem of label sparsity in the source domain. In situations where a large number of source domain samples are available but only a small portion are labeled, they leveraged cross-domain self-supervised learning to adapt the model to the target domain. Similarly, \citet{2021_WACV_GFCA} presents a method to handle the long-tailed distribution of source domain data. In cases where only a few samples are available for certain hard-to-access categories in the source domain, they improved model accuracy through generative feature augmentation.
Furthermore, \citet{2024_ICIC_PMPA_Few-Shot-Domain-Adaptation-via-Prompt-Guided} proposes a Multi-Prototype Alignment framework to transfer the model to the target domain when only a few samples per category are available in the source domain. This approach effectively addresses the scenario where there are very few samples for each category. In this paper, we tackle the challenges present in both FS-UDA and SF-UDA by training a high-performing model on the target domain despite having only a few samples per class in the source domain and no direct access to source domain samples during target domain training.


\subsection{Prompt Learning for CLIP}
In recent years, vision-language pretraining models have made significant progress in image representation learning by leveraging supervision from natural language to interpret images. Among these models, the Contrastive Language-Image Pre-training (CLIP \cite{2021_ICML_CLIP}) model is particularly representative. CLIP consists of a text encoder and an image encoder, pre-trained on 400 million image-text pairs collected from the internet, successfully aligning the feature spaces of both modalities. This approach demonstrates strong generalization capabilities in downstream tasks and enables zero-shot transfer in image classification tasks through manually specified text prompts. Inspired by prompt learning techniques in natural language processing, \citet{2022_IJCV_CoOp} introduces a learnable context optimization technique to adapt vision-language models like CLIP to image recognition tasks. Instead of relying on manually designed prompts, CoOp (Context Optimization) uses learnable vectors to model contextual words while keeping the pre-trained parameters frozen. This method allows CoOp to exhibit superior performance with limited labeled data, showing a significant improvement over manual prompt engineering in few-shot scenarios. To further enhance the alignment between visual and textual representations, \citet{2023_CVPR_MaPLe} proposes a method that applies prompt learning to both the image and text branches simultaneously, fostering a strong coupling between visual and language prompts to ensure their mutual synergy. As a result, Maple demonstrates significantly better generalization to novel classes, surpassing CoOp in performance.

\begin{figure*}[h]
    \centering
    \includegraphics[width=0.98\linewidth]{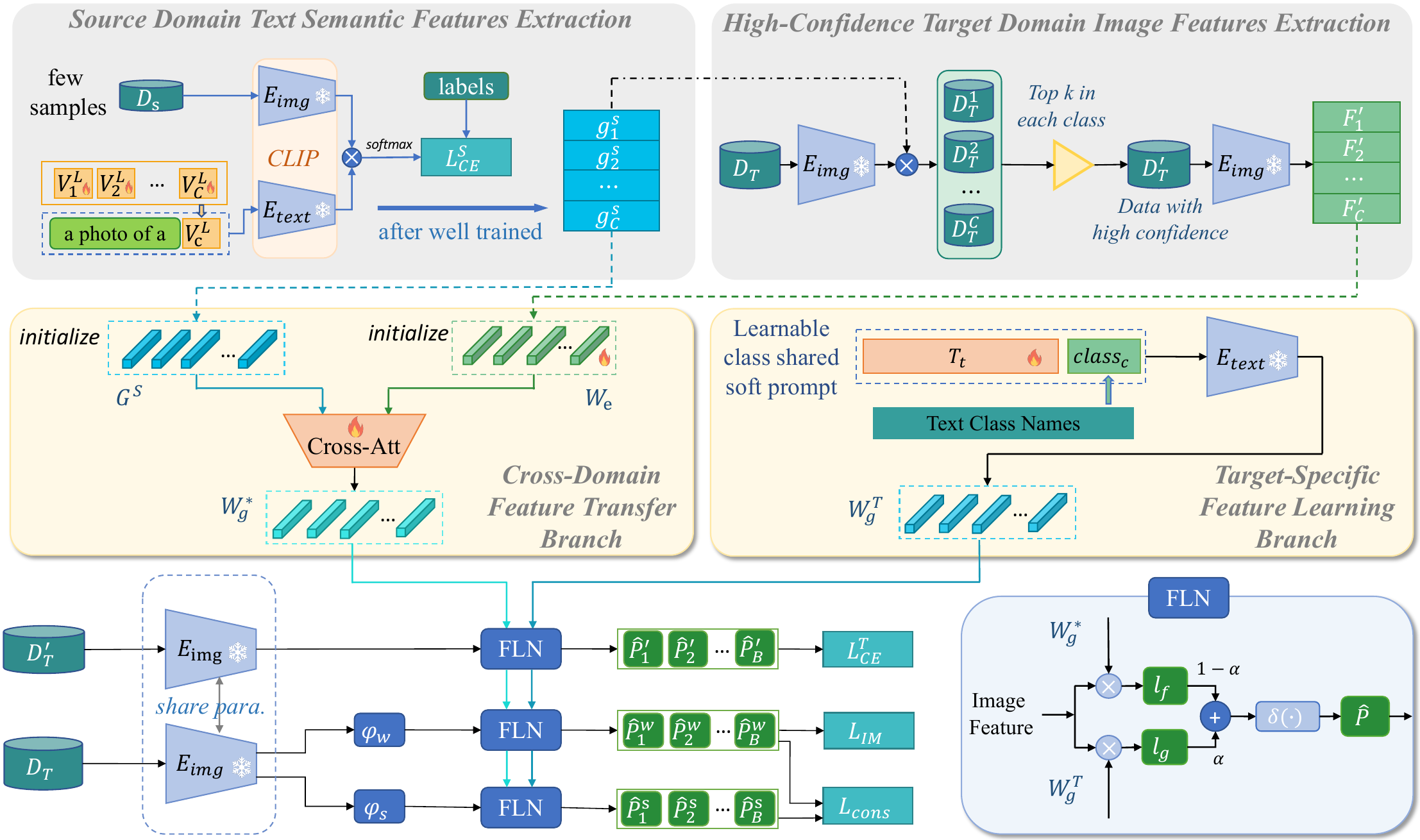}
    \caption{The framework of \ourmodel, where $E_{img}$ represents the image encoder, $E_{text}$  represents the text encoder, $T_t$ denotes the class-shared soft prompt, and $V_c^L$ indicates the learnable token for the substitute class name of class c. Solid lines represent data flow, while dashed lines indicate the flow of values without gradient back-propagation. The flame symbol represents gradient updates, and the snowflake symbol indicates frozen gradients.
    }
    \label{figure_framework}
\end{figure*}

\section{Method}
\label{sec_Method}

\subsection{Problem Definition}
Source-free unsupervised domain adaptation (SF-UDA) is an important sub-problem of unsupervised domain adaptation, focused on scenarios where source domain samples cannot be directly accessed during target domain training. This paper further explores an even more challenging scenario: the source domain samples are not only inaccessible but also limited in number.

Specifically, given a labeled source domain dataset \( D_S = \{ (x_s^i, y_s^i) \}_{i=1}^{n_s} \) with only a few samples and an unlabeled target domain dataset \( D_T = \{ x_t^i \}_{i=1}^{n_t} \), both \( D_S \) and \( D_T \) share the same labels but have different feature distributions. During the training of \( D_T \), direct access to \( D_S \) is not allowed. Instead, a model \( M_S \) is first trained using \( D_S \), and then \( M_S \) is used to train on \( D_T \). Finally, the model's performance is evaluated on \( D_T \).

\subsection{Overview of CDBN}

In this study, we aim to address the challenges of not directly accessing the source domain during target domain training, as well as the limited number of samples in the source domain. As shown in Figure 2, we propose a data-efficient CLIP-powered dual-branch network.

Specifically, we utilize the image encoder \( E_{img} \) and text encoder \( E_{text} \) from the CLIP model as the core network backbone. The model design includes a cross-domain feature transfer branch and a target-specific feature learning branch. The cross-domain feature transfer branch employs a cross-attention module with residual connections to fuse category text features \( G^s \), obtained from the source domain through prompt learning, with the features \( W_e \) of high-confidence pseudo-label samples in the target domain, facilitating feature transfer from the source domain to the target domain. The target-specific feature learning branch focuses on learning category-specific features for the target domain through a class-shared soft prompt mechanism,designed to minimize interference from noise in the source domain category text features. Finally, the model employs a fused logits network (FLN) and utilizes an unsupervised optimization strategy that combines the pseudo-label loss \( L_{CE}^T \) from high-confidence target domain samples, multi-view consistency loss \( L_{cons} \), and mutual information maximization loss \( L_{IM} \) to train the model.

\subsection{Prompt Learning in the Source Domain}

In the SF-UDA scenario with limited source domain samples, the primary challenge is how to learn semantic information from minimal data during source domain supervised training that is beneficial for unsupervised fine-tuning in the target domain. Traditional SF-UDA methods typically utilize a large number of samples to train the model \( M_S \), learning the feature distribution of the source domain. Subsequently, during the target domain training phase, information from the source domain is extracted from this model to construct a virtual domain. This surrogate aligns cross-domain feature distributions to enhance target domain performance or further fine-tunes \( M_S \) in the target domain using unsupervised or self-supervised methods to improve model performance.

However, in our scenario, due to the limited number of samples in the source domain, it is challenging to obtain a complete feature distribution through supervised learning, let alone extract the source domain feature distribution for feature alignment during target domain training. In recent years, CLIP-based prompt learning methods \cite{2022_IJCV_CoOp, 2024_AAAI_LAMM} have demonstrated the ability to learn feature representations with a certain degree of generalization from a small number of labeled samples. Inspired by these methods, we utilize a pre-trained CLIP model as the backbone of our model and learn category-specific, learnable tokens for the text encoder \( E_{text} \). This enables us to obtain category text feature representations that have generalization capabilities for domain-shared class names from the limited source domain samples.

Specifically, as shown in the Figure~\ref{figure_framework}, we freeze the parameters of CLIP and employ a commonly used textual prompt "a photo of a" with $C$ learnable class-specific tokens $V^L = \{V_0^L, V_1^L..., V_C^L\},V_c^l \in \mathcal{R}_D$, where $C$ denotes the number of classes. We concatenate the textual prompt with the class-specific tokens and use CLIP's text encoder $E_{text}$ to obtain the feature representation for each class as:
\[
g_c^S = E_{text}([[a],[photo],[of],[a],V_c^L]).
\]
For a source domain sample $x_i^S$, the probability of belonging to class $c$ is given by:

\begin{equation}
\label{eq_zero_shot}
    P( \hat{y}_i = c|{x_i^S}) = \cfrac{{\exp{\left( \frac{\left \langle E_{img}(x_i^S), g_c^S \right \rangle}{\tau} \right)}}}
    {{ \sum_{i=1}^C \exp{\left( \frac{\left \langle E_{img}(x_i^S),g_i^S \right \rangle}{\tau} \right)}}}, 
\end{equation}
where $E_{img}$ represents CLIP's image encoder, and $\tau$ is a pre-defined temperature coefficient, set to 1 by default in this work. During the training of the source domain, we optimize $V^L$ using the cross-entropy loss $\mathcal{L}_{\text{CE}}^S$ on the labeled source domain samples, as defined below:

\begin{equation}
\mathcal{L}_{\text{CE}}^S = - \frac{1}{n_s} \sum_{i=1}^{N} \sum_{c=1}^{C} y_{i,c} \log P( \hat{y}_i = c | x_i^S), 
\end{equation}
where $n_s$ denotes the number of source domain samples, and $y_{i,c}$ is the label indicating whether sample $x_i^S$ belongs to class $c$. This loss function minimizes the discrepancy between the predicted class probability distribution and the true class distribution, thereby optimizing the class-specific tokens $V^L$.

\subsection{Cross-Modal Dual-Branch Networks}
\label{section_cmdb}

After obtaining the category text features learned from the source domain data, the limited sample size and domain discrepancy lead to noise and shifts in performance within the target domain. To retain useful category semantic information for the target domain while minimizing the introduction of noise that could affect the unsupervised training of target domain samples, we designed a cross-modal dual-branch network. This network includes a cross-domain feature transfer branch to adapt and correct for domain shifts, along with a target-specific feature learning branch to optimize the feature distribution specific to the target domain, preventing the model from overfitting to the category feature distribution learned from the source domain.

First, the function of the cross-domain feature transfer branch is to receive category text features from the source domain and transfer these features to the target domain using high-confidence target domain samples. This capability arises from the alignment of image and text feature distributions during CLIP's pre-training, mapping them to a shared low-dimensional space, which allows image features to assist category text features in making predictions. Previous work \cite{2022_ECCV_Tip-Adapter} has demonstrated the feasibility of this approach by using a key-value store composed of labeled image features to enhance the classification of category text features obtained from manually designed text prompts. Inspired by earlier studies \cite{chen2024overview, 2023_NeurIPS_Meta-Adapter}, we propose to use a cross-attention module with residual connections to enable the source domain category text features to attend to the corresponding pseudo-labeled target domain image samples and facilitate their fusion, thereby transferring the source domain category text features to the target domain.

However, in our scenario, the target domain is unlabeled. To address this issue, we use the category text features from the source domain and the CLIP image encoder to obtain pseudo-labels for high-confidence samples in the target domain. As shown in the upper right corner of Figure~\ref{figure_framework}, we utilize the category feature representation $g^S$ learned from the source domain to compute the prediction results for all target domain samples using Equation~\ref{eq_zero_shot}. For each sample, the maximum category confidence is taken as its score, and the respective category is assigned as its pseudo-label. Next, based on the pseudo-labels, we partition $D_T$ into $C$ subsets $\{D_T^1,D_T^2,\dots,D_T^C \}$. For each subset, we select the $K$ samples with the highest category confidence as high-confidence samples. Finally, we aggregate the high-confidence samples from all subsets to form a highly reliable target domain sample set $D_T^{\prime}$. The reason for sorting within each subset is that the distribution of the maximum category confidence of CLIP's zero-shot is inconsistent across categories \cite{2022_arXiv_UPL}, and sorting directly among all samples would result in a long-tail distribution of selected samples.

After obtaining high-confidence target domain samples, we use the image encoder $E$ to compute features for all samples. Then, based on the samples' pseudo-labels, we construct a target domain image feature library with category information, $ F^{\prime} = [F_1^{\prime}, F_2^{\prime}, ..., F_C^{\prime}]$, where $F_c^{\prime} = [f_{0,c}^{\prime}, f_{1,c}^{\prime}, ..., f_{K,c}^{\prime}]$ and $f_{k,c}^{\prime} = E(x_k^T)$ with $x_k^T \in D_T^c$.

Then, we constructed two feature matrices: one initialized with the category text features learned from the source domain, denoted as $G^S$ with a shape of [C, D], where C is the number of categories and D is the dimension of the category text features. This matrix, containing semantic information learned from the source domain, is frozen during training. The other is a matrix $W_e$ initialized with high-confidence sample image features $F^{\prime}$ from the target domain, which has a shape of [C, K, D]. Here, $K$ represents the number of high-confidence samples per category. This matrix is modified by parameter updates during training.

Next, we use a cross-attention module with residual connections to fuse the target domain image feature matrix $W_e$ with the class text feature matrix $G^S$ learned from the source domain. 
During the fusion process, $G^S$ and $W_e$ are passed through independent linear layers and then used as the queries (Q) and keys/values (K, V) in the cross-attention layer, respectively. This approach allows each class text feature in $G^S$ to be enhanced by $K$ high-confidence samples of the corresponding class from $W_e$. By encouraging the class text features to attend to the features of different image samples in the target domain, this improves the classification performance of the class text features on target samples.

The attention output after $W_e$ and $G^S$ passes through the cross-attention layer is computed as follows:
\begin{equation}
     A = Softmax\left(\frac{(G^S \cdot W_1) \cdot (W_e \cdot W_2)^T}{\sqrt{d_k}}\right) \cdot W_e,
\end{equation}

where $W_1$ and $W_2$ are the weights of two linear layers. The weighted sum is then computed as:
\begin{equation}
    W_g^* = G^S + \alpha \cdot A,
\end{equation}
where $\alpha = W_3 \cdot G^S$ represents the class-specific weighting factor.

Finally, for a target domain sample $x_i^T$, we obtain the prediction result of this branch by calculating the cosine similarity between its image feature and the enhanced class text features as follows:
\begin{equation}
    l_f(x_i^T) =  \cfrac{W_g^* \cdot E(x_i^T)}{\|W_g^*\| \|E(x_i^T)\|} .
\end{equation}

Although the performance of the enhanced source domain class text features $W_g^*$ improves in the target domain, to further learn target-specific class text features, we introduce a branch that adapts CLIP to the target domain through learnable class-shared soft prompts, as shown in Figure~\ref{figure_framework}. Specifically, we set up a learnable sequence of class-shared tokens $T_t = [V_1^T, V_2^T, \ldots, V_M^T]$, where $V_m^T \in \mathbb{R}^D$ and $M$ are the predefined numbers of learnable tokens. During prediction, we use this token sequence to replace the original text prompt to obtain the class text features $g_c^T$. For a target domain sample $x_i^T$, the classification result of this branch is given by:

\begin{equation}
    l_g(x_i^T) =  \cfrac{W_g^T \cdot E(x_i^T)}{\|W_g^T\| \|E(x_i^T)\|} .
\end{equation}

Finally, we compute the weighted sum of the predictions from both branches to obtain the final classification result, which is formulated as follows:
\begin{equation}
\label{equ_hat_p}
    \hat{P}(x_i^T) = \delta ( \alpha l_f(x_i^T) + (1 - \alpha) l_g(x_i^T) ),
\end{equation}
where $\delta$ represents the softmax function, $\alpha$ is a predefined hyper-parameter between 0 and 1, set to 0.5 by default in this work.

\subsection{Unsupervised Optimization Strategy Driven by Accurate Classification and Diversity}
As described in Section \ref{section_cmdb}, the strategy proposed in this paper differs from most Source-Free Unsupervised Domain Adaptation (SF-UDA) methods in that it does not directly fine-tune a model trained in the source domain. Instead, we use class textual features learned from the source domain as a source of semantic information within a cross-domain feature transfer branch to balance semantic information from both the source and target domains during target domain training.
This method enables the model to effectively transfer semantic information from the source domain while minimizing the noise and domain shift effects present in the feature distribution learned from a limited number of source domain samples.
However, this strategy might result in a lack of fundamental classification capabilities at the initial stages of target domain training. To address this shortcoming, we introduce an Unsupervised Optimization Strategy Driven by Accurate Classification and Diversity, which aims to maintain the classification abilities learned from the source domain while further enhancing the model's classification performance in the target domain.

Specifically, we utilize the high-confidence sample set $D_T^{\prime}$, filtered as mentioned in Section \ref{section_cmdb}, as a reliable collection with pseudo-labels. The classification capability of the source domain model is restored through the computation of cross-entropy loss between these samples and their pseudo-labels. The formula for this computation is as follows:

\begin{equation}
\mathcal{L}_{\text{CE}}^T = -\frac{1}{B} \sum_{i=1}^{B} \sum_{c=1}^{C} \hat{y}_i^T[c] \log \hat{P}(x_i^T)[c]
\end{equation}
where $B$ is the batch size during training, $C$ is the number of classes, and $\hat{y}_i^T[k]$ is the one-hot encoding of the pseudo-label $\hat{y}_i^T$.

Although the aforementioned cross-entropy loss can enhance the model's classification capabilities, the limited number of samples involved in training results in a more uniform distribution of learned class features, which restricts its generalization ability. To address this issue, we draw inspiration from the semi-supervised learning method FixMatch \cite{2020_NeurIPS_FixMatch}, utilizing a multi-view consistency loss to increase the diversity of the class feature distribution learned by the model. This approach encourages the model to maintain consistent predictions across different augmentations of the same image, thereby broadening the feature distribution and enhancing the robustness and generalization of the model.

Specifically, for a target domain sample $x_i^T$, we apply one strong augmentation $\varphi_s$ and one weak augmentation $\varphi_w$, using the default settings specified in FixMatch. Subsequently, the predictions under these two different augmentation views, $\hat{P}_i^w$ and $\hat{P}_i^S$, are obtained as per the Equation~\ref{equ_hat_p}. For the prediction under weak augmentation, we derive its corresponding one-hot encoded pseudo-label and compute the cross-entropy loss with the prediction under strong augmentation. Furthermore, considering that pseudo-labels may contain errors, we filter out samples based on the maximum class confidence, discarding those with confidence lower than a predefined threshold $\theta_T$. The specific formula for loss calculation is as follows:

\begin{equation}
    \mathcal{L}_{\text{consistency}} = \frac{1}{B} \sum_{i=1}^{B} \mathbf{1}(\max(\hat{P}_i^W) \geq \theta_T) \cdot \text{CE}(\hat{y}_i^W, \hat{P}_i^S),
\end{equation}
where \( \mathbf{1}(\cdot) \) is an indicator function that takes the value 1 if the condition is satisfied and 0 otherwise. Here, \( \theta_T \) is set to a default value of 0.95, and \( \hat{y}_i^W = \arg\max(\hat{P}_i^W) \) is the pseudo-label generated from the predictions under weak augmentation.

Although $\mathcal{L}_{\text{consistency}}$ enhances the diversity of the model's learned class feature distribution, it fundamentally relies on pseudo-label information, which might bias the model towards categories that are easier to classify. For samples that are inherently more difficult to distinguish, the model may still make classification errors. To address this issue, and inspired by previous work \cite{2020_PMLR_SHOT, 2021_ICCV_G-SFDA}, we aim to maximize the mutual information between the model's output class distribution and the input data. This approach encourages the model's outputs to exhibit both individual certainty and global diversity, thus reducing the reliance on potentially inaccurate pseudo-labels and improving the robustness of the classification across more complex or ambiguous categories.

Specifically, this loss consists of two parts: individual entropy loss and global entropy loss. The individual entropy loss aims to reduce the classification uncertainty of each sample within a batch. The calculation formula for individual entropy loss is as follows:

\begin{equation}
    \mathcal{L}_{\text{instance}} = \frac{1}{B} \sum_{i=1}^{B} H(\mathbf{p}_i) = - \frac{1}{B} \sum_{i=1}^{B} \sum_{j=1}^{C} p_{ij} \log(p_{ij}),
\end{equation}
where \( \mathbf{p}_i \) is the predicted probability distribution for sample \( i \), and \( p_{ij} \) represents the probability of sample \( i \) belonging to class \( j \).

Global entropy loss is calculated by first determining the mean predicted probability for each class across all samples in a batch, resulting in a global probability distribution. The formula to maximize the entropy of this distribution, thus encouraging the model to encompass a broader range of classes, is as follows:

\begin{equation}
     \mathcal{L}_{\text{global}} = - \sum_{j=1}^{C} \bar{p}_j \log(\bar{p}_j),
\end{equation}
where \( \bar{p}_j \) is the average probability of samples belonging to class \( j \) across the batch.

Ultimately, the sum of the individual entropy loss and the global entropy loss serves to maximize the mutual information between the model's output class distribution and the input data. The combined loss formula is given by:
\begin{equation}
   \mathcal{L}_{\text{IM}} = - \frac{1}{N} \sum_{i=1}^{N} \sum_{j=1}^{C} p_{ij} \log(p_{ij}) + \sum_{j=1}^{C} \bar{p}_j \log(\bar{p}_j).
\end{equation}

Ultimately, during the training on the target domain, the total loss function of the model is:

\begin{equation}
   \mathcal{L}_{\text{T}} = \mathcal{L}_{\text{CE}} + \mathcal{L}_{\text{consistency}} + \mathcal{L}_{\text{IM}}.
\end{equation}

\section{Experiments}
\label{sec_Experiment}

\subsection{Experiments Setting}
\subsubsection{Datasets}

To prove the effectiveness of our proposed method, we refer to the previous method \cite{2020_PMLR_SHOT,2019_TGRS_MultiSourceRS} and select four public standard domain adaptation datasets and three publicly available remote sensing domain scene recognition datasets to evaluate our methods, which are Office-31 \cite{2010_ECCV_Adapting-Visual-Category-Models-to-New-Domains}, Office-Home \cite{2017_CVPR_Office-Home}, VisDA \cite{2017_arXiv_VisDA}, Mini-DomainNet \cite{2019_ICCV_M3SDA}, UC Merced Land-Use \cite{2010_UCM_Dataset}, NWPU-RESISC45 \cite{2017_NWPU_Dataset} and AID 
\cite{2017_AID_Dataset}. Office-31 and Office-Home are small and medium-sized datasets, VisDA and Mini-DomainNet are both large-scale datasets, and the latter two are more challenging for Source-free domain adaptation tasks. UC Merced Land-Use is a small remote sensing image dataset, while NWPU-RESISC45 and AID are two larger datasets. Overall, we validate the effectiveness of our method on these 7 datasets with 31 different transfer tasks.

\textbf{Office-31} is a small benchmark with \numComma{4652} images from 31 categories, collected from 3 different domains: Amazon (A), DSLR (D), and Webcam (W).

\textbf{Office-Home} is a medium-scale benchmark, it has \numComma{15588} images from 65 categories across 4 domains, Art(Ar), Clipart(Cl), Product(Pr), and Real-World(Rw).

\textbf{VisDA} is a large-scale dataset with synthetic source and real target domains. It has 12 classes, the synthetic source domain contains \numComma{152397} images, and the real target domain contains \numComma{55388} images.

\textbf{Mini-DomainNet} is another large-scale dataset collected from 4 domains, Real(Rl), Sketch(Sk), Clipart(Cl) and Painting(Pn). In the previous source-free domain adaptation tasks, they usually evaluate the methods on 7 transfer tasks due to the severe noisy labels in the dataset \cite{2019_ICCV_MME}. Now, we can use the clean version of the dataset and evaluate our method on 12 transfer tasks. 

\textbf{UC Merced Land-Use}, abbreviated as UCM hereafter, is a remote sensing dataset released in 2010 by the Computer Vision Laboratory at UC Merced. It is widely used for remote scene recognition to categorize land-use scenes in urban areas. The dataset contains 21 categories, each with 100 remote sensing images of size 256x256 pixels, amounting to a total of \numComma{2100} images.

\textbf{NWPU-RESISC45}, referred to as NWPU in this document, is a benchmark for remote sensing scene recognition, developed by Northwestern Polytechnical University (NWPU) in 2017. It consists of 45 scene categories, each containing 700 images, for a total of \numComma{31500} images, all with a resolution of $ 256 \times 256 $ pixels.

\textbf{AID} is a large collection of remote sensing aerial imagery. The samples are sourced from Google Earth and gathered using various remote sensing imagery sensors. The images for each category come from different countries and regions, captured at various times of the day, leading to variations in lighting conditions. Additionally, images are collected across different seasons, reflecting changes in vegetation and surface features, resulting in significant intra-class variations. The dataset consists of a total of \numComma{10000} remote sensing scene images across 30 categories, with the number of images per category ranging from 200 to 420.

\input{tabs/OfficeHome}

\input{tabs/Mini-DomainNet}

\input{tabs/Office-31}

\input{tabs/VisDA-2017}

\subsubsection{Implementation Details}
We utilized a pre-trained CLIP model as the backbone network and compared two image encoder architectures: ResNet-50\cite{he2016deep} and ViT-B/16\cite{dosovitskiy2020image}. In experiments conducted on the VisDA-2017 dataset, we employed ResNet-101 to ensure a fair comparison with other methods. The optimizer used was Adam, with a momentum of 0.9 and a weight decay set to \(5 \times 10^{-4}\). The initial learning rate was set at 0.001. During the adaptive training phase, we implemented a cosine annealing learning rate scheduler for dynamic adjustment. In our experiments, the batch size was set to 32, with $\alpha$ set to 0.5, $K$ set to 8, and $M$ set to 16. During the experiment, we randomly selected 8 samples from each category in the source domain and calculated the average of the results obtained under three different random seeds as the final result. All experiments were designed and implemented using the PyTorch platform.

\subsection{Comparison with the other methods}
\subsubsection{Competitors}
In this study, we explore a relatively under-investigated area: Source-Free Unsupervised Domain Adaptation (SF-UDA) under conditions of extremely limited source domain samples. Given the uniqueness of this novel setting, we find it challenging to identify existing studies with a completely analogous setup for comparison. Therefore, to validate the effectiveness of our method, we chose to compare it against SF-UDA methods that train with a large number of source domain samples and UDA methods that have direct access to substantial source domain samples. This comparison not only demonstrates the adaptability of our approach under the dual challenges of no direct access to source domain data and limited source sample availability but also highlights its innovation and practicality. By doing so, we aim to illustrate that even with minimal source data, our approach can still perform comparably to methods that benefit from richer source environments, highlighting its potential for scenarios where traditional UDA assumptions do not hold.

Our experimental design includes three groups of comparative methods:

\begin{itemize}
    \item  Source only, CLIP zero-shot, and Ours-Source model, which rely solely on the source domain model or zero-shot inference;
    \item  Twelve SF-UDA methods, which, although unable to directly access source domain data, can still utilize a relatively rich number of source domain samples.
    \item  Three UDA methods based on CLIP, where these approaches have direct access to the source domain data, and the source domain contains sufficient samples;
\end{itemize}

\subsubsection{Experimental Results}
First, as shown in Table~\ref{tab:OfficeHome} and Table~\ref{tab:Mini-DomainNet}, our method achieved the best performance on the standard unsupervised domain adaptation datasets Office-Home and Mini-DomainNet, with 78.1\% and 77.3\% respectively, using ResNet-50 as the backbone network. These results outperform previous source-free unsupervised domain adaptation (SF-UDA) methods as well as standard UDA approaches, clearly demonstrating that our method can achieve top performance in the target domain even under strict source data access constraints, showcasing its efficiency and robustness. Additionally, while our method did not surpass all approaches when using ViT-B/16 as the backbone, it achieved results nearly identical to the best-performing UDA method, PAD-CLIP, further validating its competitiveness.

Second, as shown in Table~\ref{tab:Office-31}, on the Office-31 dataset, our method, using ViT-B/16 as the image encoder, achieved performance comparable to SF-UDA methods based on large source samples, only 0.7\% lower than DSiT-B. Although ResNet-50 showed relatively weaker pretraining performance, our method still demonstrated its effectiveness under limited source sample conditions, proving its adaptability on datasets with smaller domain gaps.

Finally, as shown in Table~\ref{tab:VisDA-2017} and Table~\ref{tab:Recognition Results}, when using ViT-B/16 as the backbone network, our method surpassed previous SF-UDA methods on the VisDA-2017 dataset, trailing the best UDA method, PAD-CLIP, by only 1.0\%. Additionally, it achieved performance comparable to DATSNET, a UDA method specifically designed for remote sensing images, in domain adaptation tasks involving remote sensing imagery. These results not only confirm the effectiveness of our method but also highlight its significant performance advantage in professional applications with limited source samples, particularly by outperforming AD-CLIP by 3.3\%, demonstrating strong competitiveness.

These experimental results collectively validate that our proposed method can maintain strong performance across various tasks and settings, even under stringent conditions with limited source data, showcasing its robustness and generalizability.

\input{tabs/RS_res}

\input{tabs/Ablation_Sub-Components}

\subsection{Ablations Studies}

\textbf{The Effectiveness of Different Components in the Model Architecture.} In Table~\ref{tab:ablation_components}, we analyze the effectiveness of different components in the model using the Office-Home dataset, comparing the average accuracy of four distinct model configurations across the four domains. Initially, when trained solely on source domain samples, the model achieves an accuracy of 82.5\%. This indicates that prompt learning from a limited number of source domain samples retains substantial generalization information relevant to the target domain. When the model undergoes unsupervised training using only soft prompts learned from the target domain, accuracy improves to 84.3\%, confirming the effectiveness of our proposed unsupervised optimization strategy in maintaining high performance even with exclusive target domain training. Furthermore, incorporating a fusion module to integrate semantic category information from the source domain into the target domain training raises model performance to 85.7\%, demonstrating that category semantic information learned from a small number of source domain samples effectively supports training in the target domain. Finally, setting the feature matrix of high-confidence target domain samples, \(W_e\), as a learnable parameter further enhances model performance to 86.2\%. These results collectively highlight the contributions of each component to enhancing the model's generalization ability.

\begin{figure}[t]
\centering
\includegraphics[width=0.9\linewidth]{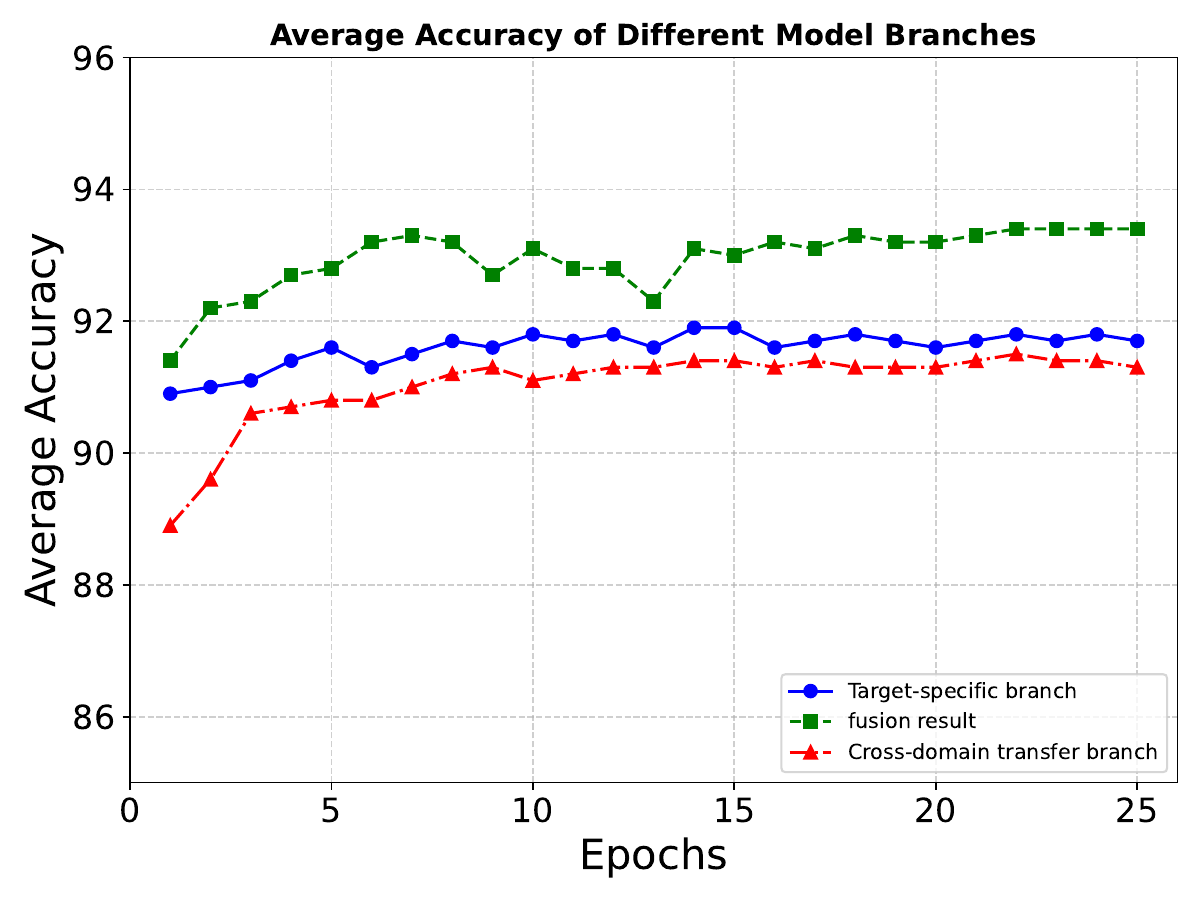} 
\caption{Trends in category average accuracy for two different branches during the Product $\to$ Real World transfer task across training epochs.}
\label{fig:dual_branch_acc}
\end{figure}

\input{tabs/Ablation_Constraint}

\textbf{The Effectiveness of Different Loss Terms in Unsupervised Optimization Strategies.} In Table~\ref{tab:ablation_losses}, we analyze the effectiveness of the three loss components in our proposed unsupervised optimization strategy across the Office-Home, VisDA-2017, and Mini-DomainNet datasets. By comparing the impact of different combinations of loss components on the accuracy of these datasets, we observe that the cross-entropy loss for high-confidence samples, \(\mathcal{L}_{CE}^T\), and the mutual information maximization loss, \(\mathcal{L}_{IM}\), significantly enhance the model's performance. This effect may stem from the fact that both \(\mathcal{L}_{CE}^T\) and \(\mathcal{L}_{IM}\) promote individual certainty in model predictions and global diversity. However, since \(\mathcal{L}_{IM}\) computes the loss over all samples, while \(\mathcal{L}_{CE}^T\) only considers a small number of highly reliable samples, the effect of \(\mathcal{L}_{IM}\) is relatively more pronounced. Furthermore, the multi-view consistency loss, \(\mathcal{L}_{\text{consistency}}\), improves model performance regardless of its combination with other losses; when all three loss components are used together, the model achieves optimal performance, indicating that each component contributes effectively.

\textbf{The Effectiveness of Logits Fusion in Dual-Branch Networks.} As shown in Figure~\ref{fig:dual_branch_acc}, we randomly selected the migration task Product → Real World from the Office-Home dataset and visualized the accuracy changes of the cross-domain transfer branch and the target-specific feature learning branch over the training epochs. The results indicate that the accuracy of both branches gradually improves as training progresses. Furthermore, the accuracy of the merged logits for both branches consistently exceeds that of any single branch in each epoch. This demonstrates the effectiveness of our proposed dual-branch logits merging strategy in enhancing model performance.

\input{tabs/Ablation_source_shots}

\begin{figure}[b]
\centering
\includegraphics[width=0.9\linewidth]{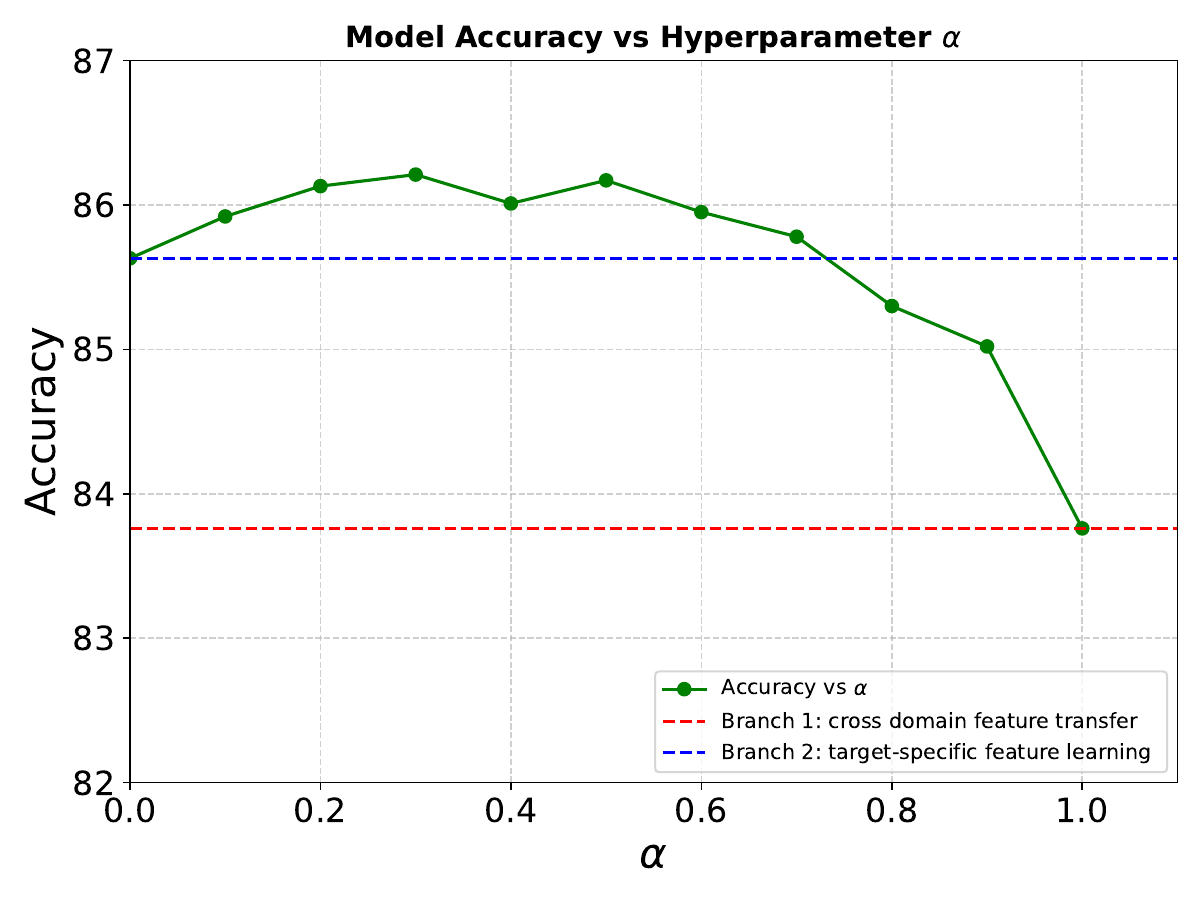} 
\caption{The average accuracy (\%) on different $\alpha$.}
\label{fig:ablation_alpha}
\end{figure}

\textbf{Impact of Source Domain Sample Size.} As illustrated in Table~\ref{tab:ablation_source_shots}, we analyzed the impact of the number of samples in the source domain on the performance of the transferred model in the target domain across three remote sensing datasets. It is evident that when the number of samples per class in the source domain is fewer than eight, increasing the sample size significantly enhances the model's performance. Specifically, the performance improved by 6.35\% when the sample count increased from one to eight per class. This improvement can be attributed to the more comprehensive semantic information about the class contained within the samples at this stage. However, when the sample count reaches sixteen per class, the model's performance does not improve and slightly decreases compared to having eight samples per class. This decline is likely due to the semantic features learned being overfitted to the source domain, and during the training of the source domain samples, the amount of learnable parameters was limited. Thus, when the number of source domain samples exceeds eight, the useful semantic information for the target domain has likely reached saturation.

\subsection{Parameter Sensitivity and Qualitative Analysis}

\textbf{The Impact of $\alpha$ on Model Performance.}
As shown in Figure~\ref{fig:ablation_alpha}, we present the trend of average accuracy across all transfer tasks in the Office-Home dataset as the parameter \(\alpha\) varies. A larger \(\alpha\) indicates a greater influence of the cross-domain feature transfer branch's predictions in the final decision. The results reveal that when \(\alpha\) is less than 0.7, the combined performance surpasses that of either branch individually. However, when \(\alpha\) exceeds 0.8, the combined performance, while still higher than that of the cross-domain transfer branch, falls below that of the target-specific feature learning branch. This is due to the lower performance of the cross-domain transfer branch compared to the target-specific branch within the Office-Home dataset. Overall, the fusion of the two branches significantly enhances their performance across most values of \(\alpha\).

\begin{figure*}[t!]
    \centering
    \subfigure[CLIP]{\includegraphics[width=0.3\textwidth]{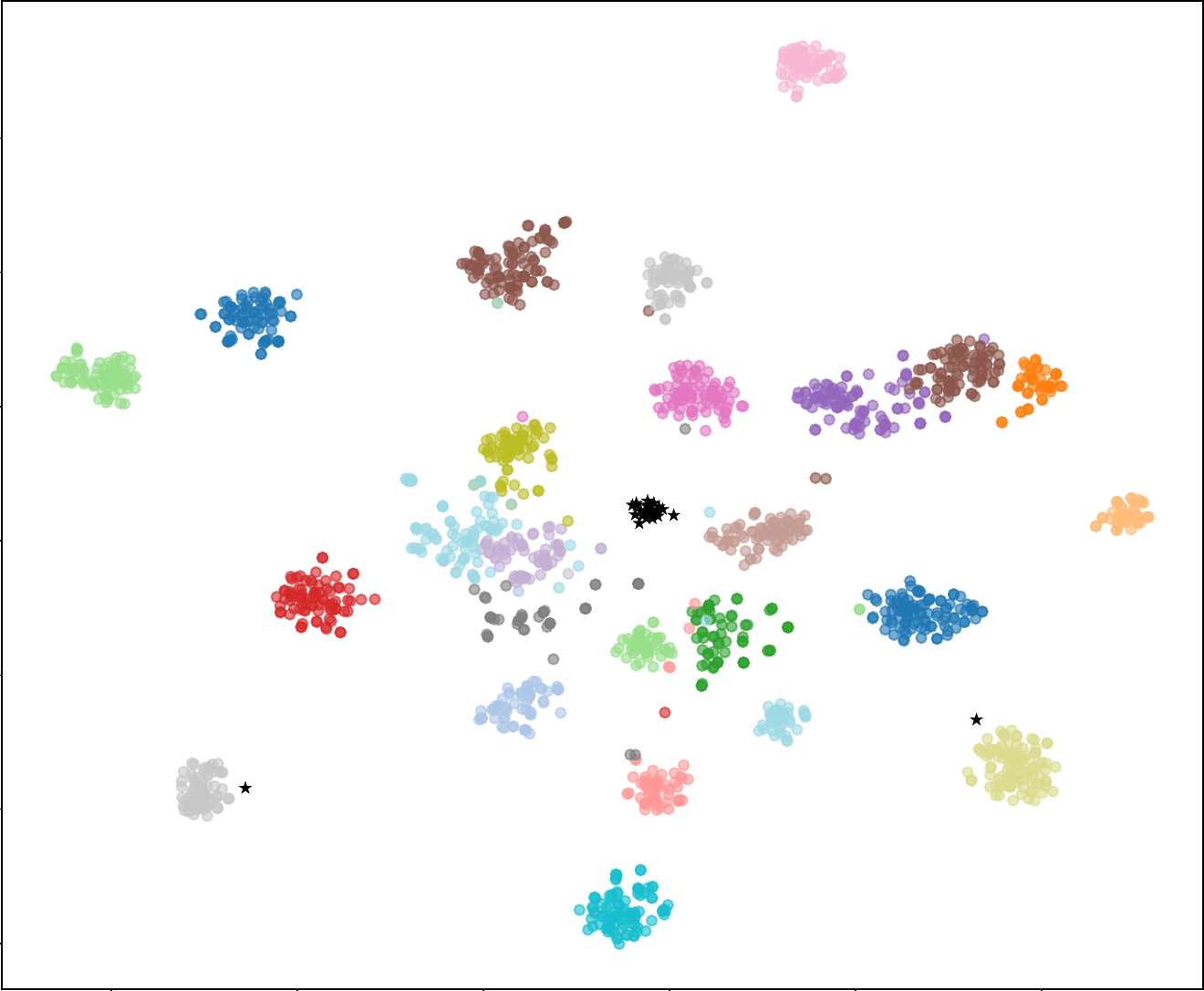}}
    \hspace{2mm}
    \subfigure[Source Only]{\includegraphics[width=0.3\textwidth]{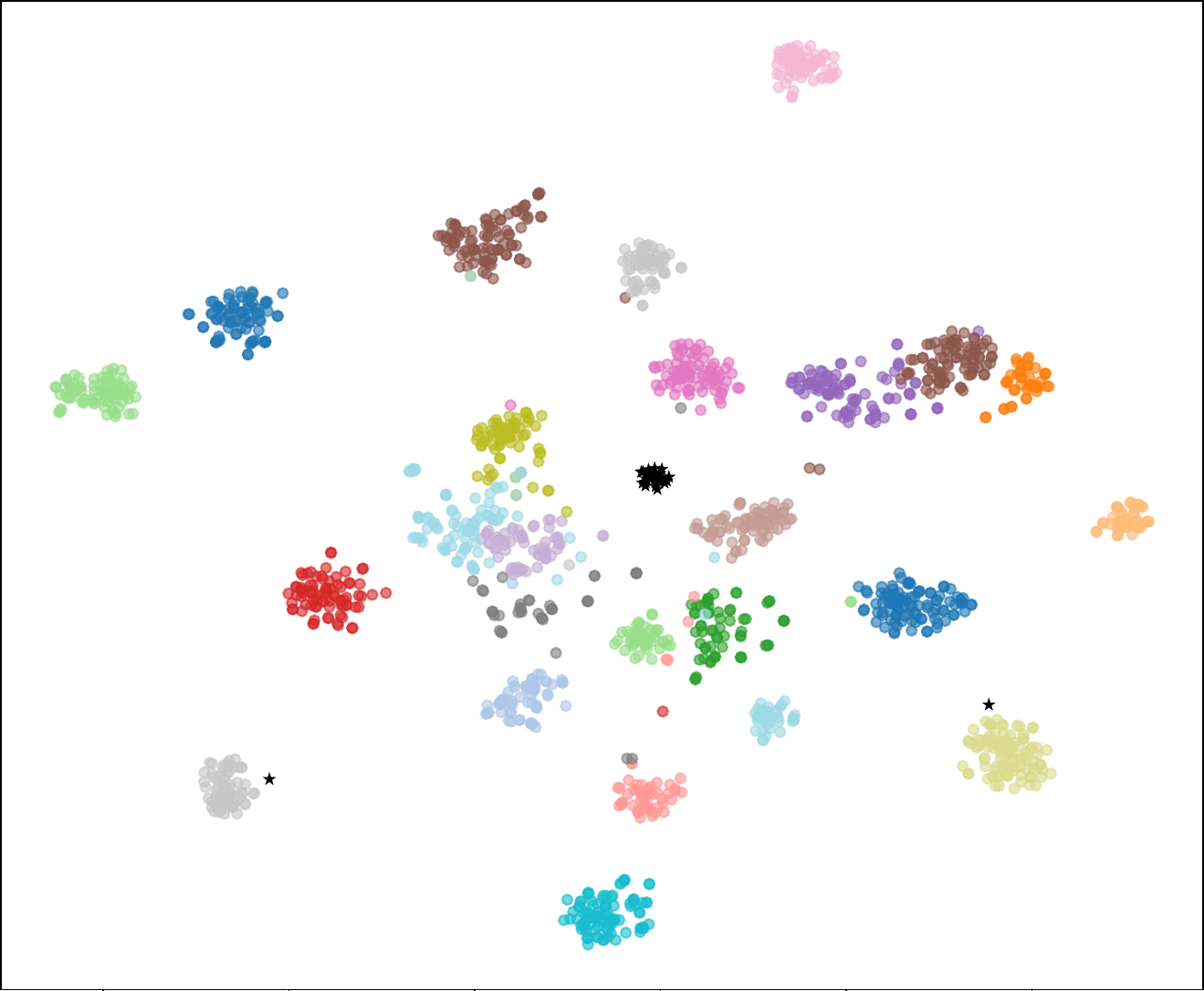}}
    \hspace{2mm}
    \subfigure[Our]{\includegraphics[width=0.3\textwidth]{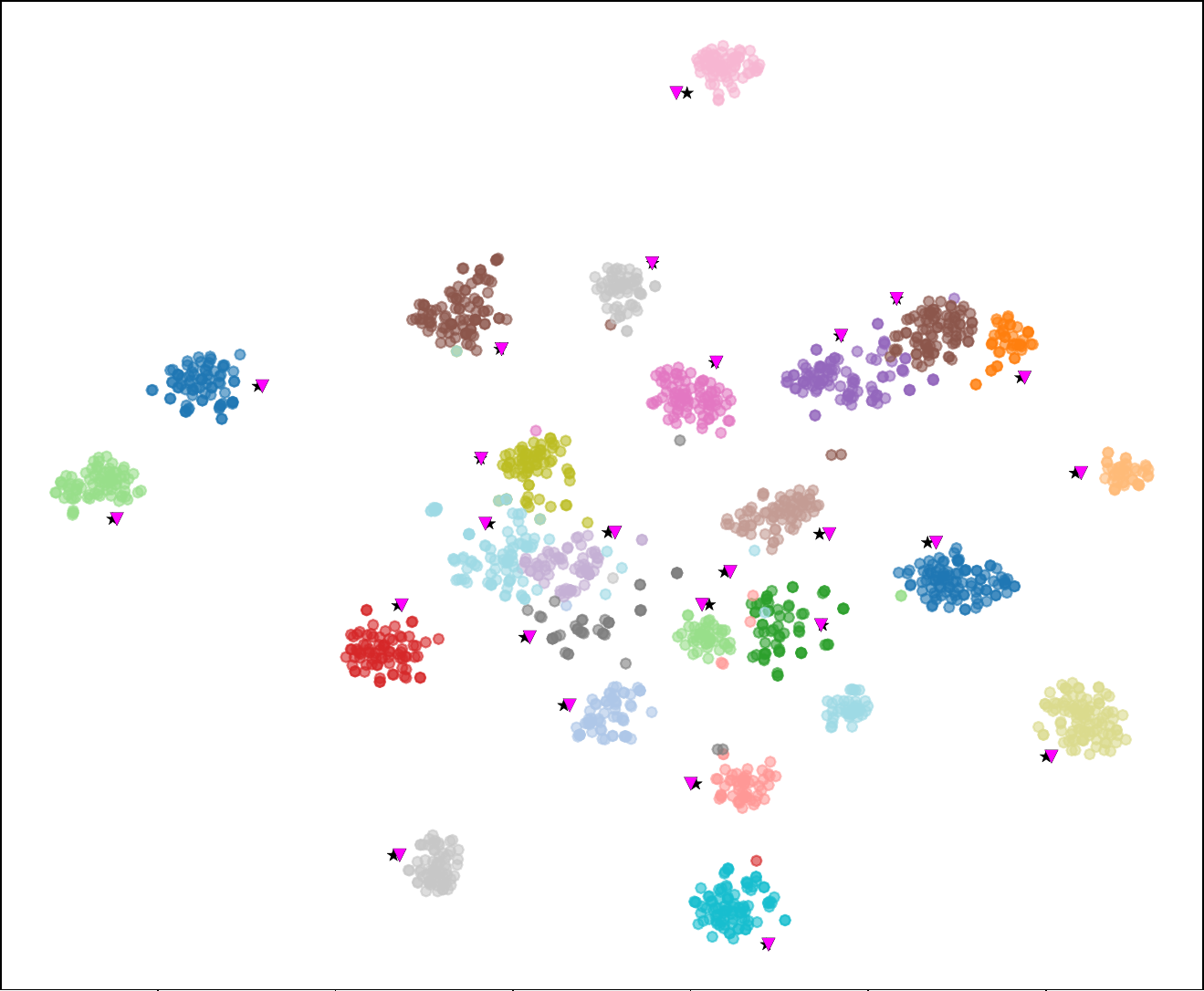}}

\caption{The t-SNE visualization of R $\to$ P transfer tasks in Office-Home. Different colored dots represent the sample features of the target domain, while black pentagrams and magenta inverted triangles denote categorical text features. (a) Categorical text features derived using the manually designed prompt "a photo of a" from CLIP; (b) Categorical text features obtained from the source domain after prompt learning; (c) Two different types of categorical text features obtained from our dual-branch network after unsupervised training, where the black pentagrams indicate the cross-domain feature transfer branch and the magenta inverted triangles indicate the target-specific feature learning branch.}
\label{fig:ablation_TSNE}
\end{figure*}

\textbf{The Impact of $K$ on Model Performance.} To analyze the impact of the number of high-confidence target domain samples within the cross-domain feature transfer branch on model performance, we conducted this experiment across 12 transfer tasks within the Office-Home dataset. We compared the average accuracy of all transfer tasks in the target domain under varying counts of high-confidence samples per category, specifically 1, 2, 4, 8, and 16 samples. As shown in Table~\ref{tab:ablation_K}, when the value of \( K \) increased from 1 to 8, there was a significant improvement in model performance, with an increase of 2.29\% when using 8 samples per category compared to just one. However, further increasing the count to 16 did not alter performance significantly. This plateau is attributed to the increased likelihood of introducing erroneous pseudo-labels as the number of samples grows, adversely affecting the model's effectiveness.

\input{tabs/Ablation_K}

\textbf{Dual-Branch Derived Category Text Features Visualization Using t-SNE.} To qualitatively analyze the variation in the class text features obtained, we conducted this experiment. Specifically, we randomly selected a transfer direction from the Office-Home dataset and, for visualization convenience, randomly chose 25 categories from the 65 available. We compared the class text features derived using the manually designed prompt "a photo of a" in CLIP, those obtained from the source domain after prompt-learning training, and those derived from our proposed dual-branch network following unsupervised training. Our method incorporates two types of class text features. As shown in Figure~\ref{fig:ablation_TSNE}, the class text features obtained from the manually designed prompt template and source domain prompt-learning are more concentrated, displaying poor class separability for the target domain samples. In contrast, after enhancement through our cross-domain transfer branch, the class text features from the source domain are dispersed across different sample clusters, and the target-specific feature learning branch also distributes class text features across various class clusters, demonstrating improved inter-class separability.

\section{Conclusion}
\label{sec_Conclusion}

In this paper, we address the dual challenges of inaccessible source domains and limited source sample availability in FS-SF-UDA by proposing a Data-Efficient CLIP-Powered Dual-Branch Network(CDBN). This network leverages the synergistic effects of cross-domain feature transfer and target-specific feature learning to enhance model generalization to the target domain while preserving the effective class semantic information learned from a small number of source domain samples. This significantly improves data utilization efficiency for source domain samples and prevents model overfitting to the limited source domain sample distribution. Furthermore, we introduce an unsupervised optimization strategy that maintains the source domain's classification capabilities while ensuring that the target domain predictions balance high accuracy with category diversity. Employing only unlabeled target domain samples, this strategy enables stable and efficient model training. Ultimately, our approach achieves best performance on multiple datasets and comparable results to the existing best methods on other datasets. This validates the effectiveness of our proposed dual-branch structure and its unsupervised optimization strategy, further demonstrating that even with limited source domain samples, the category semantic information can still provide crucial assistance for target domain adaptation.

\bibliographystyle{plainnat}

\input{main.bbl}
\vfill

\end{document}

%% file: tabs/OfficeHome.tex
\begin{table*}[!t]
\centering
\renewcommand{\arraystretch}{1.15}
\resizebox{\textwidth}{!}{
    \begin{tabular}{c|c|c|c|c|c|c|c|c|c|c|c|c|c|c|c|c}
    \toprule[1pt]
       Method & SF & Source &\textit{$E_{img}$} & Ar→Cl & Ar→Pr & Ar→Rw & Cl→Ar & Cl→Pr & Cl→Rw & Pr→Ar & Pr→Cl & Pr→Rw & Rw→Ar & Rw→Cl & Rw→Pr & Avg \\ 
    \midrule
     RN-50 \cite{2016_CVPR_ResNet} & N/A &ample& \multirow{13}{*}{\rotatebox{90}{RN-50}}&34.9 & 50.0 & 58.0 & 37.4 & 41.9 & 46.2 & 38.5 & 31.2 & 60.4 & 53.9 & 41.2 & 59.9 & 46.1 \\ 
     SHOT \cite{2020_PMLR_SHOT} & \checkmark &ample&  & 57.1 & 78.1 & 81.5 & 68.0 & 78.2 & 78.1 & 67.4 & 54.9 & 82.2 & 73.3 & 58.8 & 84.3 & 71.8 \\
     GKD \cite{2021_IROS_GKD} & \checkmark &ample&  & 56.5 & 78.2 & 81.8 & 68.7 & 78.9 & 79.1 & 67.6 & 54.8 & 82.6 & 74.4 & 58.5 & 84.8 & 72.1 \\
     PS \cite{2024_ML_PS} & \checkmark &ample&  & 57.8 & 77.3 & 81.2 & 68.4 & 76.9 & 78.1 & 67.8 & 57.3 & 82.1 & 75.2 & 59.1 & 83.4 & 72.1 \\
     D-MCD \cite{2022_AAAI_D-MCD}& \checkmark &ample&  & \textbf{59.4} & 78.9 & 80.2 & 67.2 & 79.3 & 78.6 & 65.3 & 55.6 & 82.2 & 73.3 & \textbf{62.8} & 83.9 & 72.2 \\
     A$^2$Net \cite{2021_ICCV_A2NET} & \checkmark &ample&  & 58.4 & 79.0 & 82.4 & 67.5 & 79.3 & 78.9 & 68.0 & 56.2 & 82.9 & 74.1 & 60.5 & 85.0 & 72.8 \\
     SCLM \cite{2022_NN_SCLM} & \checkmark &ample&  & 58.2 & 80.3 & 81.5 & 69.3 & 79.0 & 80.7 & 69.0 & 56.8 & 82.7 & 74.7 & 60.6 & 85.0 & 73.1 \\
     AAA \cite{2021_PAMI_AAA} & \checkmark &ample&  & 56.7 & 78.3 & 82.1 & 66.4 & 78.5 & 79.4 & 67.6 & 53.5 & 81.6 & 74.5 & 58.4 & 84.1 & 71.8 \\
     CoWA \cite{2022_PMLR_CoWA} & \checkmark &ample&  & 56.9 & 78.4 & 81.0 & 69.1 & 80.0 & 79.9 & 67.7 & 57.2 & 82.4 & 72.8 & 60.5 & 84.5 & 72.5 \\
     ProxyMix \cite{2023_NN_ProxyMix} &\checkmark & ample& &  \underline{59.3} & 81.0 & 81.6 & 65.8 & 79.7 & 78.1 & 67.0 & 57.5 & 82.7 & 73.1 & \underline{61.7} & 85.6 & 72.8\\
     C-SFDA \cite{2023_ICCV_C-SFDA} & \checkmark &ample&  & 60.3 & 80.2 & 82.9 & 69.3 & 80.1 & 78.8 & 67.3 & 58.1 & 83.4 & 73.6 & 61.3 & 86.3 & 73.5 \\
     TPDS \cite{2024_IJCV_TPDS} &\checkmark & ample&  &  59.3 & 80.3 & 82.1 & 70.6 & 79.4 & 80.9 & 69.8 & 56.8 & 82.1 & 74.5 & 61.2 & 85.3 & 73.5 \\
     CPD \cite{2024_PR_CPD} &\checkmark & ample&  &  59.1 & 79.0 & 82.4 & 68.5 & 79.7 & 79.5 & 67.9 & 57.9 & 82.8 & 73.8 & 61.2 & 84.6 & 73.0 \\
     \midrule
     CLIP \cite{2021_ICML_CLIP} & N/A & no & \multirow{6}{*}{\rotatebox{90}{RN-50}}& 51.6 & 81.9 & 82.6 & 71.9 & 81.9 & 82.6 & 71.9 & 51.6 & 82.6 & 71.9 & 51.6 & 81.9 & 72.0 \\
     DAPL \cite{2023_TNNLS_DAPL} & \texttimes &ample&  & 54.1 & 84.3 & 84.8 & 74.4 & 83.7 & 85.0 & 74.5 & 54.6 & 84.8 & 75.2 & 54.7 & 83.8 & 74.5 \\
     PADCLIP \cite{2023_ICCV_padclip} & \texttimes &ample&  & 57.5 & 84.0 & 83.8 & \textbf{77.8} & 85.5 & 84.7 & 76.3 & \textbf{59.2} & \underline{85.4} & \textbf{78.1} & 60.2 & \underline{86.7} & \underline{76.6}\\
     ADCLIP \cite{2023_ICCV_ADCLIP} &   \texttimes &ample&  &55.4 & \underline{85.2} & \underline{85.6} & 76.1 & \underline{85.8} & \underline{86.2} & \textbf{76.7} & 56.1 & \underline{85.4} & 76.8 & 56.1 & 85.5 & 75.9\\
     Ours-Source & N/A & 8-shots&  &52.7 & 84.8 & 83.6 & 70.2 & 83.2 & 82.6 & 69.7 & 50.3 & 82.8 & 71.6 & 51.6 & 84.3 & 72.2\\
      \cellcolor{lightblue!30} Ours & \checkmark & 8-shots &  & 59.2 & \textbf{89.4} & \textbf{87.1} & \underline{77.0} & \textbf{89.0} & \textbf{87.0} & \underline{76.6} & \underline{59.1} & \textbf{87.3} & \underline{76.9} & 59.9 & \textbf{88.9} &  \cellcolor{lightblue!30}\textbf{78.1}\\
    \midrule
     DSiT-B* \cite{2023_ICCV_DSiT-B} & \checkmark &ample&  \multirow{7}{*}{\rotatebox{90}{ViT-B/16}} & 69.2 & 83.5 & 87.3 & 80.7 & 86.1 & 86.2 & 77.9 & 67.9 & 86.6 & 82.4 & 68.3 & 89.8 & 80.5\\
     CLIP \cite{2021_ICML_CLIP} & N/A &no&  & 67.8 & 89.0 & 89.8 & 82.9 & 89.0 & 89.8 & 82.9 & 67.8 & 89.8 & 82.9 & 67.8 & 89.0 & 82.4 \\
     DAPL \cite{2023_TNNLS_DAPL} &\texttimes &ample&  & 70.7 & 91.0 & 90.9 & 85.2 & 91.0 & 91.0 & 85.1 & 70.7 & 90.9 & 85.3 & 70.4 & 91.4 & 84.4 \\
     PADCLIP \cite{2023_ICCV_padclip} & \texttimes &ample&  & \textbf{76.4} & 90.6 & 90.8 & \textbf{86.7} & 92.3 & \underline{92.0} & \underline{86.0} & \textbf{74.5} & 91.5 & \textbf{86.9} & \textbf{79.1} & 93.1 & \textbf{86.7}\\
     ADCLIP \cite{2023_ICCV_ADCLIP} & \texttimes &ample&  & 70.9 & \underline{92.5} & \textbf{92.1} & 85.4 & \underline{92.4} & \textbf{92.5} & \textbf{86.7} & \underline{74.3} & \textbf{93.0} & \textbf{86.9} & 72.6 & \underline{93.8} & 86.1\\
     Ours-Source & N/A & 8-shots& & 69.0 & 89.6 & 90.1 & 81.8 & 91.1 & 89.9 & 79.7 & 67.8 & 88.9 & 82.5 & 68.6 & 90.4 & 82.5\\
     \cellcolor{lightblue!30} Ours & \checkmark & 8-shots&  & \underline{73.9} & \textbf{94.1} & \underline{91.6} & \underline{85.7} & \textbf{94.3} & 91.7 & 84.9 & 73.7 & \underline{91.9} & \underline{85.5} & \underline{72.7} & \textbf{94.0} & \cellcolor{lightblue!30}\underline{86.2} \\ 
    \bottomrule
    \end{tabular}
}
    \caption{Accuracy (\%) of Different Settings and Domain Adaptation Methods on the Office-Home Dataset \cite{2017_CVPR_Office-Home}. SF indicates the accessibility of the source domain, Source refers to the number of samples in the source domain, and $E_{img}$ denotes the image encoder used. The highest accuracy for each encoder is highlighted in bold, while the second highest is underlined for clarity.}
    
    \label{tab:OfficeHome}
\end{table*}

%% file: tabs/Mini-DomainNet.tex
\begin{table*}[!t]
\centering
\renewcommand{\arraystretch}{1.15}
\resizebox{\textwidth}{!}{
    \begin{tabular}{c|c|c|c|c|c|c|c|c|c|c|c|c|c|c|c|c}
    \toprule[1pt]
       Method & SF &Source & $E_{img} $ & Cl→Pn & Cl→Rl & Cl→Sk & Pn→Cl & Pn→Rl & Pn→Sk & Rl→Cl & Rl→Pn & Rl→Sk & Sk→Cl & Sk→Pn & Sk→Rl & Avg \\ 
    \midrule
     CLIP \cite {2021_ICML_CLIP} & N/A & no & \multirow{5}{*}{\rotatebox{90}{RN-50}} & 67.9 & 84.8 & 62.9 & 69.1 & 84.8 & 62.9 & 69.2 & 67.9 & 62.9 & 69.1 & 67.9 & 84.8 & 71.2 \\ 
     DAPL \cite{2023_TNNLS_DAPL} & \texttimes& ample& &\underline{72.4} & 87.6 & 65.9 & 72.7 & \underline{87.6} & \underline{65.6} & 73.2 & \underline{72.4} & 66.2 & \underline{73.8} & 72.9 & 87.8 & 74.8\\
     ADCLIP \cite{2023_ICCV_ADCLIP} & \texttimes& ample& &71.7 & \underline{88.1} & \underline{66.0} & \underline{73.2} & 86.9 & 65.2 & \underline{73.6} & \textbf{73.0} & \underline{68.4} & 72.3 & \underline{74.2} & \underline{89.3} & \underline{75.2} \\
     Ours-Source & N/A & 8-shots & &71.0 & 87.3 & 63.7 & 70.8 & 86.0 & 61.1 & 68.6 & 68.1 & 60.5 & 71.1 & 70.3 & 85.7 & 72.0 \\
     \cellcolor{lightblue!30} Ours &  \checkmark&  8-shots & &\textbf{75.4} & \textbf{89.2} & \textbf{67.3} &\textbf{77.5} & \textbf{89.5} & \textbf{67.9} & \textbf{76.8} & 68.1 & \textbf{77.0} & \textbf{74.9} & \textbf{74.8} & \textbf{89.4} & \cellcolor{lightblue!30}\textbf{77.3} \\
     \midrule 
     CLIP \cite{2021_ICML_CLIP} &N/A & no & \multirow{5}{*}{\rotatebox{90}{ViT-B/16}} &80.3 & 90.5 & 77.8 & 82.7 & 90.5 & 77.8 & 82.7 & 80.3 & 77.8 & 82.7 & 80.3 & 90.5 & 82.8 \\
     DAPL \cite{2023_TNNLS_DAPL} & \texttimes& ample& &83.3 & 92.4 & 81.1 & 86.4 & 92.1 & 81.0 & 86.7 & 83.3 & \underline{80.8} & \underline{86.8} & \underline{83.5} & \underline{91.9} & 85.8 \\
     ADCLIP \cite{2023_ICCV_ADCLIP} & \texttimes&ample&  &\underline{84.3} & \textbf{93.7} & \textbf{82.4} & \underline{87.5} & \textbf{93.5} & \textbf{82.4} & \underline{87.3} & \underline{84.5} & \textbf{81.6} & \textbf{87.9} & \textbf{84.8} & \textbf{93.0} & \textbf{86.9} \\
     Ours-Source & N/A &8-shots&  &81.0 & 90.6 & 79.2 & 85.9 & 90.3 & 78.1 & 84.3 & 78.1 & 77.6 & 84.8 & 80.8 & 90.5 & 83.4 \\
     \cellcolor{lightblue!30} Ours &  \checkmark&  8-shots & & \textbf{84.8} & \underline{92.4}	& \underline{81.9} & \textbf{87.6} & \underline{93.0} & \underline{82.1} & \textbf{87.5} & \textbf{84.9} & \textbf{81.6} & \textbf{87.9} & \textbf{84.8}& \underline{91.9}& \cellcolor{lightblue!30}\underline{86.7} \\
    \bottomrule[1pt]
    \end{tabular}

}
    \caption{Accuracy (\%) of Different Settings and Domain Adaptation Methods on the Mini-DomainNet\cite{2019_ICCV_M3SDA} dataset.  }
    
    \label{tab:Mini-DomainNet}
\end{table*}

%% file: tabs/Office-31.tex
\begin{table}[!t]
    \renewcommand{\arraystretch}{1.6}
    \centering
    \scalebox{0.66}{
        \begin{tabular}{c|c|c|c|c|c|c|c|c|c|c}
            \toprule[1pt]
            Method & SF & Source & $E_{img} $ & A→D & A→W & D→A & D→W & W→A & W→D & Avg. \\ 
            \midrule
            RN-50 \cite{2016_CVPR_ResNet} & N/A &ample&  \multirow{14}{*}{\rotatebox{90}{RN-50}} & 68.9 & 68.4 & 62.5 & 96.7 & 60.7 & 99.3 & 76.1 \\
            SHOT \cite{2020_PMLR_SHOT} & \checkmark & ample& &  94.0 & 90.1 & 74.7 & 98.4 & 74.3 & \underline{99.9} & 88.6\\
            GKD \cite{2021_IROS_GKD} &\checkmark &ample&  &  94.6 & 91.6 & 75.1 & 98.7 & 75.1 & \textbf{100.0} & 89.2\\
            D-MCD \cite{2022_AAAI_D-MCD} &\checkmark &ample&  & 94.1 & 93.5 & 76.4 & \underline{98.8} & 76.4 & \textbf{100.0} & 89.9  \\
            A$^2$Net \cite{2021_ICCV_A2NET} &\checkmark &ample&  & 94.5 & 94.0 & 76.7 & 99.2 & 76.1 & \textbf{100.0} & 90.0  \\
            SCLM \cite{2022_NN_SCLM} &\checkmark & ample& &  95.8 & 90.0 & 75.5 & \textbf{98.9} & 75.5 & 99.8 & 89.4 \\
            AAA \cite{2021_PAMI_AAA} & \checkmark &ample&  &  95.6 & 94.2 & 75.6 & 98.1 & 76.0 & 99.8 & 89.9 \\ 
            CoWA \cite{2022_PMLR_CoWA} & \checkmark & ample& &  94.4 & \underline{95.2} & 76.2 & 98.5 & 77.6 & 99.8 & 90.3 \\ 
            ProxyMix \cite{2023_NN_ProxyMix} &\checkmark &ample&  & 95.4 & \textbf{96.6} & 75.1 & 98.5 & 75.4 & 99.8 & 85.6 \\
            C-SFDA \cite{2023_ICCV_C-SFDA} &\checkmark &ample&  & 96.2 & 93.9 & \underline{77.3} & \underline{98.8} & 77.9 & 99.7 & \underline{90.5} \\
            TPDS \cite{2024_IJCV_TPDS} &\checkmark &ample&  & \textbf{97.1} & 94.5 & 75.7 & 98.7 & 75.5 & 99.8 & 90.2 \\
            CPD \cite{2024_PR_CPD} &\checkmark & ample& & \underline{96.6} & 94.2 & \underline{77.3} & 98.2 & \underline{78.3} & \textbf{100.0} & \textbf{90.8}\\
            CLIP \cite{2021_ICML_CLIP} &N/A & no & & 74.3 & 67.8 & 72.6 & 67.8 & 72.6 & 74.3 & 71.6 \\
            Ours-Source &N/A & 8-shots & & 78.9 & 70.7 & 74.2 & 83.0 & 72.1 & 82.3 & 76.9 \\
            \cellcolor{lightblue!30}Ours &\checkmark &  8-shots & & 90.0 & 88.6 &\textbf{80.3} & 91.1 & \textbf{79.6} & 92.4 & \cellcolor{lightblue!30}87.0 \\
            \midrule 
            DSiT-B* \cite{2023_ICCV_DSiT-B} &\checkmark &ample&  \multirow{3}{*}{\rotatebox{90}{\tiny ViT-B/16}} & \textbf{98.0} & \textbf{97.2} & \underline{81.7} & \textbf{99.1} & \underline{81.8} & \textbf{100.0} & \textbf{93.0} \\
            CLIP \cite{2021_ICML_CLIP} &N/A & no & & 79.9 & 76.9 & 78.9 & 76.9 & 78.9 & 79.9 & 78.6 \\
            Ours-Source &N/A& 8-shots & & 83.3 & 81.5 & 79.7 & 90.1 & 78.7 & 91.6 & 84.2 \\
            \cellcolor{lightblue!30}Ours &\checkmark &  8-shots & & \underline{94.8} & \underline{95.8} & \textbf{85.1} & \underline{95.8} & \textbf{85.6} & \underline{96.6} & \cellcolor{lightblue!30}\underline{92.3} \\
            \bottomrule[1pt]
        \end{tabular}
    }
    \caption{Accuracy (\%) of Different Settings and Domain Adaptation Methods on the Office-31\cite{2010_ECCV_Adapting-Visual-Category-Models-to-New-Domains} Dataset. }
    \label{tab:Office-31}
\end{table}

%% file: tabs/VisDA-2017.tex
\begin{table*}[t]
\centering
    \resizebox{0.99\linewidth}{!}{
    \fontsize{6}{7}\selectfont
    \begin{tabular}{c|c|c|c|c|c|c|c|c|c|c|c|c|c|c|c|c}
        \toprule[1pt]
        Method & SF &Source & $E_{img}$ & plane & bcycl & bus & car & horse & knife & mcycle & person & plant & sktbrd & train & truck & Avg. \\ 
        \midrule
        RN-101 \cite{2016_CVPR_ResNet} & N/A & ample &\multirow{13}{*}{\rotatebox{90}{RN-101}}  & 55.1 & 53.3 & 61.9 & 59.1 & 80.6 & 17.9 & 79.7 & 31.2 & 81.0 & 26.5 & 73.5 & 8.5 & 52.4 \\
        SHOT \cite{2020_PMLR_SHOT} & \checkmark &ample& & 94.3 & 88.5 & 80.1 & 57.3 & 93.1 & 94.9 & 80.7 & 80.3 & 91.5 & 89.1 & 86.3 & 58.2 & 82.9 \\
        GKD \cite{2021_IROS_GKD} & \checkmark & ample& & 95.3 & 87.6 & 81.7 & 58.1 & 93.9 & 94.0 & 80.0 & 80.0 & 91.2 & 91.0 & 86.9 & 56.1 & 83.0 \\
        PS \cite{2024_ML_PS} & \checkmark& ample& & 95.2 & 86.2 & 82.1 & 61.5 & 93.2 & 95.6 & 86.7 & 80.4 & 91.5 & 90.8 & 85.8 & 59.3 & 84.0 \\
        D-MCD \cite{2022_AAAI_D-MCD} &\checkmark & ample& & 97.0 & 88.0 & 90.0 & 81.5 & 95.6 & \textbf{98.0} & 86.2 & \textbf{88.7} & 94.6 & 92.7 & 83.7 & 53.1 & 87.5 \\
        A$^2$Net \cite{2021_ICCV_A2NET} &\checkmark &ample&  & 94.0 & 87.8 & 85.6 & 66.8 & 93.7 & 95.1 & 85.8 & 81.2 & 91.6 & 88.2 & 86.5 & 56.0 & 84.3 \\
        SCLM \cite{2022_NN_SCLM} &\checkmark &ample&  & 97.1 & 90.7 & 85.6 & 62.0 & 97.3 & 94.6 & 81.8 & 84.3 & 93.6 & 92.8 & 88.0 & 55.9 & 85.3\\
        AAA \cite{2021_PAMI_AAA}&\checkmark &ample& & 94.4 & 85.9 & 74.9 & 60.2 & 96.0 & 93.5 & 87.8 & 80.8 & 90.2 & 92.0 & 86.6 & \underline{68.3} & 84.2\\
        CoWA \cite{2022_PMLR_CoWA} &\checkmark &ample& & 96.2 & 89.7 & 83.9 & 73.8 & 96.4 & \underline{97.4} & 89.3 & \underline{86.8} & 94.6 & 92.1 & 88.7 & 53.8 & 86.9\\
        ProxyMix \cite{2023_NN_ProxyMix} &\checkmark &ample& & 95.4 & 81.7 & 87.2 & 79.9 & 95.6 & 96.8 & 92.1 & 85.1 & 93.4 & 90.3 & 89.1 & 42.2 & 85.7\\ 
        C-SFDA \cite{2023_ICCV_C-SFDA} &\checkmark & ample& & 97.6 & 88.8 & 86.1 & 72.2 & 97.2 & 94.4 & 92.1 & 84.7 & 93.0 & 90.7 & \textbf{93.1} & 63.5 & \underline{87.8} \\ 
        TPDS \cite{2024_IJCV_TPDS} &\checkmark & ample& & 97.6 & \textbf{91.5} & 89.7 & \textbf{83.4} & \underline{97.5} & 96.3 & 92.2 & 82.4 & \textbf{96.0} & \textbf{94.1} & 90.9 & 40.4 & 87.6\\ 
        CPD \cite{2024_PR_CPD} &\checkmark & ample& & 96.7 & 88.5 & 79.6 & 69.0 & 95.9 & 96.3 & 87.3 & 83.3 & 94.4 & 92.9 & 87.0 & 58.7 & 85.8\\ 
        \midrule
        CLIP \cite{2021_ICML_CLIP} & N/A & no & \multirow{6}{*}{\rotatebox{90}{RN-101}} &\textbf{98.2} & 83.9 & 90.5 & 73.5 & 97.2 & 84.0 & 95.3 & 65.7 & 79.4 & 89.9 & 91.8 & 63.3 & 84.4 \\
        DAPL \cite{2023_TNNLS_DAPL} &\texttimes & ample& & 97.8 & 83.1 & 88.8 & 77.9 & 97.4 & 91.5 & 94.2 & 79.7 & 88.6 & 89.3 & 92.5 & 62.0 & 86.9 \\
        ADCLIP \cite{2023_ICCV_ADCLIP} &\texttimes & ample& & \underline{98.1} & 83.6 & \underline{91.2} & 76.6 & \textbf{98.1} & 93.4 & \textbf{96.0} & 81.4 & 86.4 & 91.5 & 92.1 & 64.2 & 87.7\\ 
        PADCLIP \cite{2023_ICCV_padclip} &\texttimes & ample& & 96.7 & 88.8 & 87.0 & \underline{82.8} & 97.1 & 93.0 & 91.3 & 83.0 & \underline{95.5} & 91.8 & 91.5 & 63.0 & \textbf{88.5}\\ 
        Ours-Source & N/A &8-shots&  & 96.2 & 76.3 & \textbf{95.4} & 66.5 & 97.1 & 62.6 & \underline{95.6} & 9.3 & 92.4 & \underline{93.3} & 91.5 & 43.0 & 76.6 \\ 
        \cellcolor{lightblue!30} Ours &\checkmark &  8-shots & & 97.3 & \underline{90.8} & 83.5 & 67.9 & 96.8 & 95.0 & 90.4 & 78.5 & 89.3 & 91.4 & \underline{92.8} & \textbf{72.5} & \cellcolor{lightblue!30}87.2 \\
        \midrule
        DSiT-B* \cite{2023_ICCV_DSiT-B} &\checkmark &ample&  \multirow{7}{*}{\rotatebox{90}{ViT-B/16}} & - & - & - & - & - & - & - & - & - & - & - & - & 87.6 \\ 
        CLIP \cite{2021_ICML_CLIP} & N/A & no &  & \underline{99.1} & 91.7 & 93.8 & 76.7 & \underline{98.4} & 91.7 & \underline{95.3} & 82.7 & 86.5 & 96.0 & \underline{94.6} & 60.5 & 88.9\\ 
        DAPL \cite{2023_TNNLS_DAPL} &\texttimes &ample&  & - & - & - & - & - & - & - & - & - & - & - & - & 89.8 \\
        ADCLIP \cite{2023_ICCV_ADCLIP} &\texttimes & ample& & \textbf{99.6} & 92.8 & \underline{94.0} & \underline{78.6} & \textbf{98.8} & \underline{95.4} & \textbf{96.8} & \underline{83.9} & \underline{91.5} & 95.8 & \textbf{95.5} & 65.7 & \underline{90.7}\\ 
        PADCLIP \cite{2023_ICCV_padclip} &\texttimes & ample& & 98.1 & \underline{93.8} & 87.1 & \textbf{85.5} & 98.0 & \textbf{96.0} & 94.4 & \textbf{86.0} & \textbf{94.9} & 93.3 & 93.5 & \underline{70.2} & \textbf{90.9}\\  
        Ours-Source & N/A & 8-shots & & \underline{99.1} & 89.7 & \textbf{94.1} & 75.1 & 98.1 & 55.1 & 93.3 & 78.7 & 86.8 & \textbf{96.7} & 94.5 & 51.9 & 84.4\\ 
        \cellcolor{lightblue!30} Ours &\checkmark &  8-shots & & 97.9 & \textbf{94.0} & 88.1 & 70.3 & \underline{98.4} & 95.0 & 93.4 & 82.5 & 91.1 & 95.9 & 92.9 & \textbf{78.9} & \cellcolor{lightblue!30}89.9 \\
        \bottomrule[1pt]
    \end{tabular}
}
\caption{Accuracy (\%) of Different Settings and Domain Adaptation Methods on the VisDA-2017\cite{2017_arXiv_VisDA} Dataset.}
\label{tab:VisDA-2017}
\end{table*}

%% file: tabs/RS_res.tex
\begin{table}[!t]
\centering
\renewcommand{\arraystretch}{1.6}
\scalebox{0.62}{
        \begin{tabular}{c|c|c|c|c|c|c|c|c|c|c}
    \toprule[1pt]
    Method & SF & Source & $f_v$ & A$\to$U & U$\to$A & N$\to$U & U$\to$N & N$\to$A & A$\to$N & Average \\
    \midrule
    RN-50 \cite{2016_CVPR_ResNet} & N/A & ample & \multirow{7}{*}{\rotatebox{90}{RN-50}}  & 62.3 & 61.8 & 70.2 & 61.6 & 88.2 & 74.5 & 69.8 \\
    DDC \cite{2014_arxiv_DDC}  & \texttimes & ample& & 68.4 & 52.8 & 80.5 & 52.1 & 85.3 & 81.3 & 70.1 \\
    DAN \cite{2017_TGRS_DAN}  &\texttimes  & ample &  &  75.4 & 56.6 & 85.3 & 54.2 & 86.7 & 86.7 & 74.1 \\
    Deep CORAL \cite{2016_ECCV_DeepCoral} &\texttimes  &ample & &  65.7 & 53.9 & 79.8 & 52.2 & 85.1 & 81.4 & 69.7 \\
    DANN  \cite{2016_JMLR_DANN}  & \texttimes & ample & &  67.2 & 52.8& 84.4 & 53.3 & 86.1 & 82.3 & 71.0 \\
    DATSNET \cite{2022_TGRS_DATSNET} & \texttimes &ample & &  85.7 & 88.4 & 91.9 & 77.5 & 94.3 & 88.4 & 87.7 \\
    \midrule
    CLIP \cite{2021_ICML_CLIP} & N/A & no &\multirow{3}{*}{\rotatebox{90}{RN-50}}  & 54.7& 63.4& 49.2& 49.7& 56.8& 48.2& 53.7\\
    Ours-Source & \texttimes&8-shots & &  71.0 & 75.1 & 65.2	& 60.8& 81.8& 70.8	&70.8 \\
    \cellcolor{lightblue!30}Ours& \checkmark & no & & 88.7& 93.8 &	85.5 & 75.5 & 92.2& 84.1& \cellcolor{lightblue!30}86.6 \\
    \midrule
    CLIP \cite{2021_ICML_CLIP} & N/A & no &\multirow{5}{*}{\rotatebox{90}{ViT-b/16}}   & 67.8	& 70.9 & 62.5 & 61.4 & 62.6 &  56.2 & 63.6 \\
    DAPL \cite{2023_TNNLS_DAPL} &\texttimes & ample &&	71.4 &  78.1 & 64.7 & 64.9 & 65.8 & 61.7 & 67.8 \\
    AD-CLIP \cite{2023_ICCV_ADCLIP} & \texttimes & ample& & 86.9 & 95.4 & 91.2 & 87.7  &93.9 & 87.9 & 90.5 \\
    Ours-Source  & N/A & 8-shots  & &  81.9 & 92.0 & 81.5 & 84.1 & 87.6 & 82.9 & 85.0 \\
    \cellcolor{lightblue!30}Ours& \checkmark& no & & \textbf{92.3} & \textbf{98.8} & \textbf{92.0} & \textbf{91.6} & \textbf{96.3} & \textbf{91.5} & \cellcolor{lightblue!30}\textbf{93.8} \\
    \bottomrule[1pt]
    \end{tabular}
}
\caption{Accuracy (\%) for cross-domain remote scene recognition on the UCM\cite{2010_UCM_Dataset}, AID\cite{2017_AID_Dataset}, and NWPU\cite{2017_NWPU_Dataset} datasets. }
\label{tab:Recognition Results}
\end{table}

%% file: tabs/Ablation_Sub-Components.tex
\begin{table}[b]
    \centering
    \renewcommand{\arraystretch}{1.45}
    \resizebox{\linewidth}{!}{
    \begin{tabular}{ccccc}
    \toprule
        Source training & Target Prompt & Fusion dual branches & Trainable $W_e$ & Avg\\
        \midrule
        \checkmark & & & & 82.5\\
        & \checkmark & & & 84.3\\
        \checkmark& \checkmark & \checkmark & & 85.7\\
        \checkmark& \checkmark& \checkmark & \checkmark &\textbf{86.2}\\
        \bottomrule
    \end{tabular}
    }
    \caption{Ablation study of sub-components of the proposed method measured by classification accuracy(\%) on Office-Home benchmark, 8 shots for cache key, backbone ViT-B/16. }
    \label{tab:ablation_components}
\end{table}

%% file: tabs/Ablation_Constraint.tex
\begin{table}[h]
    \centering
    \renewcommand{\arraystretch}{1.4}
    \resizebox{\linewidth}{!}{
    \begin{tabular}{ccccccc}
    \toprule
        $\mathcal{L}_{ce}$& $\mathcal{L}_{im}$& $\mathcal{L}_{fm}$& Office-Home & DomainNet-126 & VisDA-2017 & Avg\\
        \midrule
        \checkmark & & & 84.2& 84.4& 89 & 86.17\\
        & \checkmark & & 85.2& 86.2& 89.3& 86.90\\
        \checkmark & & \checkmark & 85.0& 86& 89.9& 86.97\\
 \checkmark& \checkmark& & 85.6& 85.8&89.3& 86.90\\
        & \checkmark & \checkmark & 85.4& 86.1& 89.4& 86.97\\
        \checkmark & \checkmark& \checkmark & \textbf{86.2}& \textbf{86.7}& \textbf{89.9} & \textbf{87.60}\\
        \bottomrule
    \end{tabular}
    }
    \caption{Ablation on different constraint losses, backbone ViT-B/16.}
    \label{tab:ablation_losses}
\end{table}

%% file: tabs/Ablation_source_shots.tex
\begin{table}[h]
    \centering
    \renewcommand{\arraystretch}{1.6}
    \resizebox{\linewidth}{!}{
    \begin{tabular}{cccccccc}
    \toprule[1pt]
        shots & A $\to$ U & U $\to$ A & N $\to$ U & U $\to$ N & N $\to$ A & A $\to$ N & Average \\ 
        \hline
        1 & 82.1 & 98.2 & 82.1 & 83.6 & 87.6 & 90.8 & 87.40  \\ 
        2 & 86.8 & \textbf{98.9} & 81.6 & 91.2 & 95 & 88.5 & 90.33  \\ 
        4 & 88.1 & 98.8 & 91.0 & 91.2 & \textbf{98.9} & 90.1 & 93.02  \\ 
        8 & \textbf{92.3} & 98.8 & \textbf{92.0} & 91.6 & 96.3 & \textbf{91.5} & \textbf{93.75}  \\ 
        16 & 92.1 & 98.6 & 91.2 & \textbf{92.1} & 96.1 & 91.0 & 93.52 \\ 
    \bottomrule[1pt]
    \end{tabular}
    }
    \caption{Model performance in target domain with different source domain sample sizes.}
    \label{tab:ablation_source_shots}
\end{table}

%% file: tabs/Ablation_K.tex
\begin{table}[h]
    \centering
    \resizebox{0.9\linewidth}{!}{
    \fontsize{4.5}{5}\selectfont
    \renewcommand{\arraystretch}{1.1}
    \begin{tabular}{cccccc}
        \toprule[0.6pt]
        K & 1 & 2 & 4 & 8 & 16 \\
        \midrule
        Avg & 75.83 & 76.6 & 77.34 & \textbf{78.12} & 78.08 \\
        \bottomrule[0.6pt]
    \end{tabular}
    }
    \caption{Average Accuracy for Different Values of $K$ on the OfficeHome Dataset Using ResNet-50 Backbone.}
    \label{tab:ablation_K}
\end{table}